\documentclass[10pt]{article}
\usepackage[preprint]{tmlr}

\usepackage{amsmath,amsfonts,bm}

\def\eqref#1{equation~\ref{#1}}

\def\1{\bm{1}}

\DeclareMathAlphabet{\mathsfit}{\encodingdefault}{\sfdefault}{m}{sl}
\SetMathAlphabet{\mathsfit}{bold}{\encodingdefault}{\sfdefault}{bx}{n}

\usepackage{hyperref}
\usepackage{url}
\usepackage{fontawesome5}
\usepackage{tikz}
\definecolor{HFYellow}{HTML}{FFD21E}
\newcommand{\hficon}{%
  \begin{tikzpicture}[baseline=-0.62ex,x=0.66em,y=0.66em,line cap=round]
    \fill[HFYellow] (0,0) circle[radius=0.62];
    \fill (-0.18,0.13) circle[radius=0.075];
    \fill ( 0.18,0.13) circle[radius=0.075];
    \draw[line width=0.055em] (-0.22,-0.13) .. controls (-0.07,-0.32) and (0.07,-0.32) .. (0.22,-0.13);
  \end{tikzpicture}%
}
\newcommand{\vriqiconlink}[2]{{\hypersetup{hidelinks}\href{#1}{\textcolor{black}{#2}}}}
\newcommand{\linkVRIQ}{\vriqiconlink{https://huggingface.co/datasets/tina-khezresmaeilzadeh/VRIQ}{\hficon\,VRIQ}}
\newcommand{\linkDiag}{\vriqiconlink{https://huggingface.co/datasets/tina-khezresmaeilzadeh/DiagVRIQ}{\hficon\,DiagVRIQ}}
\newcommand{\linkCode}{\vriqiconlink{https://github.com/tinakhezresmaeilzadeh/vriq}{\faGithub\,Code}}
\usepackage{amsmath,amssymb}
\usepackage{booktabs}
\usepackage{multirow}
\usepackage{array}
\usepackage{makecell}
\usepackage{graphicx}
\usepackage[dvipsnames]{xcolor}
\usepackage[most]{tcolorbox}
\usepackage{enumitem}
\usepackage{caption}
\usepackage{tabularx}

\newtcolorbox{llmresponse}{colback=gray!5,colframe=gray!60,boxrule=0.5pt,
  arc=1mm,left=4pt,right=4pt,top=3pt,bottom=3pt,fontupper=\small}

\title{VRIQ: Benchmarking and Diagnosing the Visual-Reasoning IQ of Vision--Language Models}

\author{%
  \name Tina Khezresmaeilzadeh\textsuperscript{1}\thanks{Corresponding author: \href{mailto:khezresm@usc.edu}{khezresm@usc.edu}}
  \quad Jike Zhong\textsuperscript{1}
  \quad Konstantinos Psounis\textsuperscript{1}\\
  {\normalfont\normalsize\textsuperscript{1}University of Southern California}\\[0.45em]
  {\normalfont\normalsize \linkVRIQ\quad\linkDiag\quad\linkCode}
}

\def\month{MM}
\def\year{YYYY}
\def\openreview{\url{https://openreview.net/forum?id=XXXX}}

\begin{document}
\maketitle

\begin{abstract}
Vision--language model (VLM) errors on visual reasoning may come from
missing visual structure, from mistakes after that structure is
extracted, or from both. We introduce \textbf{VRIQ}, a benchmark of 1{,}391
human-verified IQ-style visual reasoning puzzles spanning five
psychometric categories, each instantiated in category-matched
abstract and natural-image domains. To study perception and reasoning
more directly, we also introduce \textbf{DiagVRIQ}, a 209-item
diagnostic subset paired with human-verified, rule-neutral textual
descriptions. Across 21 VLMs, the 17 non-frontier models reach a 26.8\%
unweighted category mean on abstract puzzles, while three 2026
frontier models improve substantially but unevenly, with unweighted
category means of 41.3--70.9\% on abstract puzzles and at most 78.2\%
on natural-image puzzles; all remain below human performance. On DiagVRIQ,
even when their descriptions are judged correct,
the strongest models still answer approximately 9--12\% of those
items incorrectly. Two interventions, ground-truth description
augmentation and tool-augmented inference, improve accuracy, with the
largest tool gains on spatially demanding categories, but neither
eliminates errors. \textbf{VRIQ} and \textbf{DiagVRIQ} provide a
framework for studying visual reasoning through complementary
end-to-end, description-based, and intervention-based analyses rather
than treating it as a single monolithic capability.
\end{abstract}

\section{Introduction}
\label{sec:intro}

Recent years have witnessed remarkable advances in both Large Language
Models (LLMs) and Vision--Language Models (VLMs), driven by new
architectures~\citep{vaswani2017attention,brown2020language,touvron2023llama},
improved training methodologies~\citep{chung2024scaling,ouyang2022training},
and large-scale datasets~\citep{schuhmann2022laion,gadre2023datacomp}.
Structured inference methods such as chain-of-thought
prompting~\citep{wei2022chain}, self-consistency~\citep{wang2023selfconsistency},
and tree-of-thoughts~\citep{yao2023tree} have made reasoning a central
capability of modern language models.

Extending these capabilities to the visual domain has proven harder.
Visual reasoning requires models both to extract relevant information
from an image and to apply relational, spatial, or abstract operations
over it, abilities that matter as VLMs are deployed in settings such as
medical diagnosis~\citep{li2023llavamed,phan2024decomposing} and
autonomous navigation~\citep{pan2024vlp,chen2024asynchronous}. Despite
work on visual chain-of-thought~\citep{shao2024visual}, tool-augmented
reasoning~\citep{gupta2023visual,suris2023vipergpt}, and models that
actively inspect or manipulate images during
inference~\citep{openai-thinking-images,zheng2026deepeyes}, recent
benchmarks continue to reveal substantial gaps in vision-centric and
abstract visual reasoning~\citep{xu2026visulogic,song2025visualpuzzles}.

A central difficulty is that end-to-end accuracy does not reveal where
a failure arises. A model may fail to extract relevant visual
structure, fail to reason over information it has extracted, or both.
Distinguishing these cases matters because they suggest different
remedies, from stronger visual representations to improved inference
procedures.

Recent benchmarks increasingly target abstract, vision-centric, or
knowledge-light visual reasoning~\citep{jiang2024marvel,
xu2026visulogic,song2025visualpuzzles}, while established IQ-style
benchmarks evaluate abstract reasoning without a corresponding
natural-image domain~\citep{zhang2019raven,barrett2018measuring}.
Diagnostic benchmarks that separate perception from reasoning either
progressively supply ground-truth information that eventually reveals
the rule~\citep{chia2024puzzlevqa} or pose separate perception
questions~\citep{jiang2024marvel}. These approaches do not jointly
measure, item by item, whether a model's own description of the visual
content is correct and whether it then answers the puzzle correctly.

\begin{figure}[htbp]
    \centering
    \includegraphics[width=0.85\textwidth]{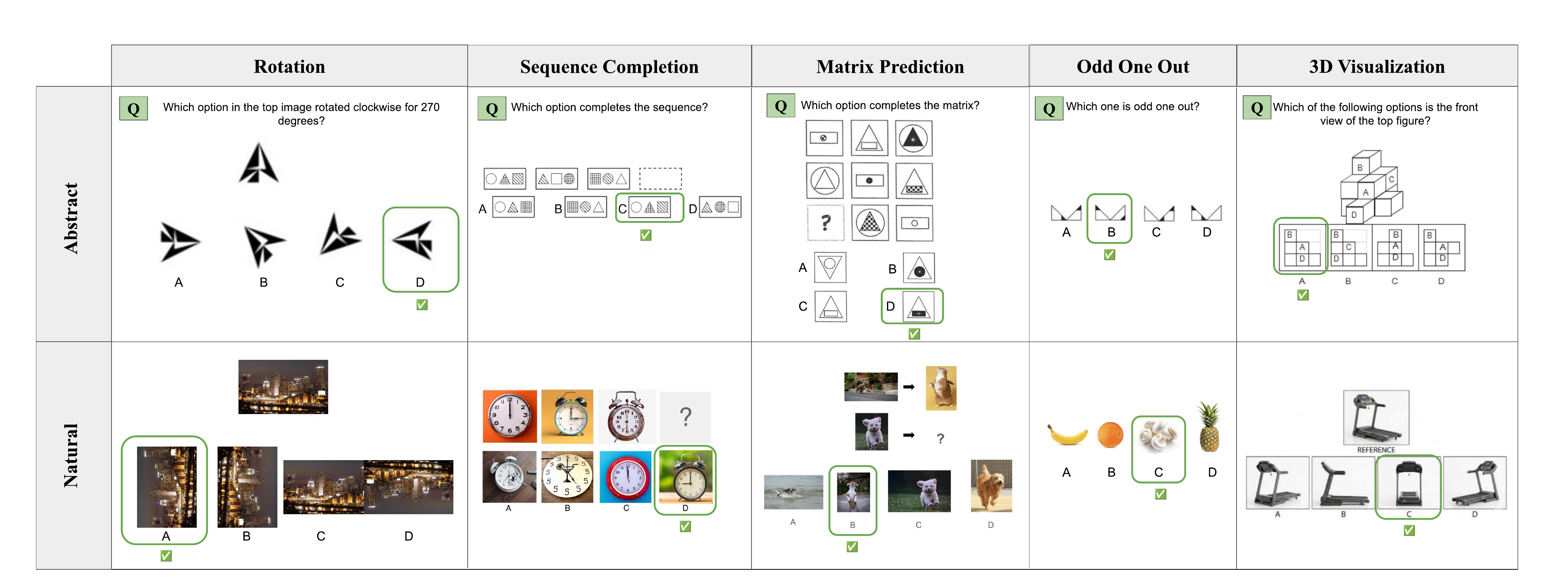}
    \caption{VRIQ overview. Five reasoning categories are instantiated
    in abstract and natural-image domains following the same category
    definitions. The domains are matched at the category level rather
    than item by item.}
    \label{fig:overview}
\end{figure}

To address this gap, we introduce \textbf{VRIQ}, a benchmark of
1{,}391 human-verified IQ-style visual reasoning puzzles spanning five
categories drawn from nonverbal aptitude testing
(\autoref{fig:overview}). Each category is instantiated in an abstract
and a natural-image domain following the same category definition; the
domains are matched at the category level rather than item by item.

To study perception and reasoning more directly, we construct
\textbf{DiagVRIQ}, a 209-item subset of natural-domain Matrix
Prediction, Odd One Out, and Sequence Completion puzzles whose visual
content can be faithfully represented in text without revealing the
rule or answer. Each item is paired with an LLM-drafted,
human-verified, rule-neutral reference description. In a
describe-then-solve protocol, each model produces a description and an
answer in the same response, so description correctness and answer
correctness can be analyzed on the same items. We calibrate the
description judge against human annotation on all five evaluated
models and check it with an independent cross-family judge. We
complement this analysis with two interventions:
\emph{ground-truth description augmentation}, which supplies verified
visual information without revealing the rule or answer, and
\emph{tool-augmented inference}, which lets a reasoning model crop,
zoom, rotate, and otherwise re-inspect images.

Across 21 VLMs, including three 2026 frontier systems, visual-reasoning
performance improves substantially at the frontier, but unevenly across
models, domains, and categories. On DiagVRIQ, even the strongest models
answer approximately 9--12\% of items incorrectly despite descriptions
judged correct.

In summary, our contributions are:
\begin{itemize}[leftmargin=*,itemsep=2pt,topsep=2pt]

    \item \textbf{VRIQ and DiagVRIQ}: a 1{,}391-item benchmark of
    category-matched abstract and natural-image reasoning tasks,
    together with a 209-item diagnostic subset containing verified,
    rule-neutral descriptions for item-level analysis of description
    correctness and answer correctness.

    \item \textbf{A description-based diagnostic protocol} that
    evaluates each model's own visual description against a verified
    reference and measures answer accuracy conditional on description
    correctness, with a judge calibrated against human annotation and
    checked by a cross-family judge.

    \item \textbf{Two targeted interventions}, ground-truth
    description augmentation and tool-augmented inference, together
    with an evaluation spanning open-weight, proprietary, and 2026
    frontier models, including a sensitivity analysis of
    inference-time reasoning effort.

\end{itemize}

VRIQ, DiagVRIQ, ground-truth descriptions, and evaluation code are publicly available
(see links on the title page).

\section{Related Work}
\label{sec:related}

\paragraph{Multimodal benchmarks.}
General-purpose visual question answering (VQA) benchmarks cover
natural images, documents, scene text, and knowledge-intensive
questions~\citep{antol2015vqa,goyal2017making,hudson2019gqa,
marino2019ok,singh2019towards,gurari2018vizwiz,mathew2021docvqa,
li2024seed,liu2023mmbench,yue2024mmmu}, while specialized benchmarks
target mathematical, scientific, chart, and engineering
reasoning~\citep{lu2021inter,lu2023mathvista,zhang2024mathverse,
wang2024measuringmultimodalmathematicalreasoning,qiao2024we,
liang2024scemqa,wang2023scibench,chen2023theoremqa,wang2024charxiv,
eeebench,doris2024designqamultimodalbenchmarkevaluating}. These
benchmarks primarily evaluate end-to-end task performance.
MathVerse~\citep{zhang2024mathverse} is a notable exception that varies
how much problem information is carried by the diagram versus the text,
but it targets mathematical diagrams rather than nonverbal reasoning.

\paragraph{Abstract and IQ-style visual reasoning.}
Procedurally generated datasets such as
PGM~\citep{barrett2018measuring} and RAVEN~\citep{zhang2019raven}
established abstract visual reasoning as a testbed for relational
inference, primarily for models trained on the task itself. More recent
benchmarks evaluate general-purpose VLMs zero-shot.
PuzzleBench~\citep{zhang2025puzzlebenchfullydynamicevaluation} uses
dynamically generated puzzles to limit contamination;
VisualPuzzles~\citep{song2025visualpuzzles} targets knowledge-light
algorithmic, analogical, deductive, inductive, and spatial reasoning;
and VisuLogic~\citep{xu2026visulogic} emphasizes vision-centric
reasoning designed to reduce language-based shortcuts. MM-IQ
~\citep{cai2025mm} evaluates VLMs on a large collection of IQ-test
items. VRIQ differs in instantiating each category in both an abstract
and a natural-image domain and in providing item-level diagnostic
analysis.

\paragraph{Diagnosing perception versus reasoning.}
PuzzleVQA~\citep{chia2024puzzlevqa} diagnoses models by progressively
supplying ground-truth explanations of visual content, inductive
reasoning, and deductive reasoning. Because later stages reveal the
rule, this protocol measures how much each hint helps rather than
whether the model's own visual account was correct.
MARVEL~\citep{jiang2024marvel} instead pairs abstract puzzles with
separately posed perception questions. DiagVRIQ differs by evaluating
each model's own description against a verified, rule-neutral
reference and conditioning answer accuracy on description correctness
within the same model output. Individual components have precedents;
our contribution is their integration with category-matched domains
and two interventions in a single benchmark
(\autoref{tab:related-comparison}).

\paragraph{Tool-augmented visual reasoning.}
Visual programming approaches decompose visual tasks into explicit
operations or invoke external vision
tools~\citep{gupta2023visual,suris2023vipergpt}, while visual
chain-of-thought encourages intermediate reasoning over image
regions~\citep{shao2024visual}. Recent systems also support or learn
active visual inspection during reasoning
~\citep{openai-thinking-images,zheng2026deepeyes}. We use tool access
as an intervention on the same benchmark: the same model is evaluated
with and without tools, with image manipulation, additional
computation, and code execution treated as a bundled setting.

\begin{table*}[t]
\centering
\footnotesize
\caption{Comparison with selected closely related abstract visual
reasoning benchmarks. VRIQ's natural-image domain is matched to the
abstract domain by category, not item by item.}
\label{tab:related-comparison}

\setlength{\tabcolsep}{3pt}
\renewcommand{\arraystretch}{1.15}

\begin{tabularx}{\textwidth}{
l
c
c
>{\raggedright\arraybackslash}X
>{\raggedright\arraybackslash}X
>{\raggedright\arraybackslash}X
}
\toprule
Benchmark &
\makecell{Abstract\\items} &
\makecell{Natural-image\\domain} &
How perception is assessed &
Perceptual information supplied &
Interventions \\
\midrule

PGM / RAVEN
& \checkmark
& --
& Not assessed
& None
& -- \\

MM-IQ
& \checkmark
& --
& Not assessed
& None
& -- \\

PuzzleVQA
& \checkmark
& --
& Staged diagnostic hints
& Explanations including the rule
& Staged hints \\

MARVEL
& \checkmark
& --
& Separate perception questions
& None
& -- \\

\midrule

\textbf{VRIQ (ours)}
& \checkmark
& \makecell{\checkmark\\category-matched}
& Model-generated descriptions
& Verified rule-neutral descriptions
& \makecell[l]{GT-description augment\\and tool-based inference} \\

\bottomrule
\end{tabularx}
\end{table*}
\section{VRIQ: A Diagnostic Visual Reasoning Benchmark}
\label{sec:benchmark}

VRIQ (Visual Reasoning IQ) is a benchmark of 1{,}391 human-verified
visual reasoning puzzles designed to evaluate vision--language models
on tasks requiring both visual information extraction and abstract
reasoning. The benchmark spans five psychometric reasoning categories:
3D Visualization (3D), Figure Rotation (FR), Matrix Prediction (MP),
Odd One Out (OO), and Sequence Completion (SC). Each category is
instantiated in two category-matched domains: an \emph{abstract}
domain built from geometric primitives and a \emph{natural} domain
built from real-world objects and scenes.
\autoref{tab:dataset-stats} gives the composition.

\begin{table}[t]
\centering
\small
\caption{VRIQ composition by category and domain. DiagVRIQ items are
a subset of the natural domain.}
\label{tab:dataset-stats}
\begin{tabular}{lcccccc}
\toprule
Domain & 3D & FR & MP & OO & SC & Total \\
\midrule
Abstract          & 210 & 198 & 141 & 106 & 133 & 788 \\
Natural           & 111 &  97 & 146 & 149 & 100 & 603 \\
\midrule
Total             & 321 & 295 & 287 & 255 & 233 & 1{,}391 \\
\midrule
DiagVRIQ (natural subset) & -- & -- & 98 & 51 & 60 & 209 \\
\bottomrule
\end{tabular}
\end{table}

\subsection{Benchmark Construction}
\label{sec:construction}

\paragraph{Abstract-domain puzzles.}
Abstract puzzles consist of geometric shapes, patterns, and spatial
transformations with carefully designed distractors. Abstract items
come from public aptitude and civil-service materials, reformatted and
modified in shape, count, orientation, or styling while keeping the
rule, and from puzzles written by the authors. The 3D items were built
in Tinkercad~\citep{tinkercad}. Figure Rotation questions were
generated by Python code. Details are provided in
\autoref{app:abstract-sourcing}.

\paragraph{Natural-domain puzzles.}
Natural puzzles instantiate the same category-level reasoning
requirements using real-world objects and scenes. The questions were
written by the authors. Images were either generated with a
commercial image model or drawn from publicly available,
copyright-compliant sources (generators in
\autoref{app:generation-procedure}). Human annotators selected,
edited, and verified every retained item.

\paragraph{Quality control.}
All puzzles were independently solved by two annotators and retained
only when both agreed on a unique correct answer. Natural-domain
items were additionally reviewed for rule consistency, visual clarity,
and the absence of shortcut cues, such as an answer option identifiable
without applying the intended reasoning rule.

\paragraph{Contamination mitigation.}
Because some abstract puzzles originate from publicly accessible
examination materials, training-data contamination cannot be fully
ruled out. Adapted abstract items were modified in shape, count,
orientation, or styling while keeping the rule, and other abstract
items, including the Python-generated Figure Rotation questions, were
written by the authors. Natural-domain questions were written for
this benchmark. Their images were either generated or taken from public
sources, and the assembled puzzles have no public counterparts. These
measures reduce
but do not fully eliminate the possibility of contamination,
particularly for the abstract domain. All description-based attribution
and ground-truth augmentation experiments are conducted on newly
authored natural-domain items.

\subsection{Diagnostic Annotations}
\label{sec:diagnostic-annotations}

VRIQ includes \textbf{DiagVRIQ}, a diagnostic subset designed for
item-level analysis of the relationship between visual description
correctness and reasoning performance.

\paragraph{DiagVRIQ: a subset for perception--reasoning analysis.}
Matrix Prediction, Odd One Out, and Sequence Completion contain
natural-domain items whose visual content can be faithfully represented
in text without revealing the intended reasoning rule. We screened
items in these three categories for this property and retained those
for which annotators could construct a complete textual description
without substantial loss of task-relevant visual information and
without stating or implying the rule or correct answer. This procedure
yielded 209 items: 98 Matrix Prediction, 51 Odd One Out, and
60 Sequence Completion items.

Each retained item is paired with a verified ground-truth description.
Descriptions were initially drafted by GPT-5.2~\citep{chatgpt-5.2}
and then verified and, where necessary, revised by human annotators to
ensure coverage of the relevant panels, entities, attributes, and
answer options. Each description is restricted to observable visual
content and is both \emph{rule-neutral} and \emph{answer-neutral}: it
neither states the underlying reasoning rule nor identifies or
privileges a particular answer option. The authoring and verification
protocol is given in \autoref{app:gt-description-protocol}.

These annotations support both the describe-then-solve analysis
(\autoref{sec:description-attribution-method}) and the ground-truth
description augmentation experiment
(\autoref{sec:reasoning-with-gt}). 3D Visualization and Figure
Rotation are excluded from DiagVRIQ because their dense spatial
structure could not be faithfully serialized into text without
substantial information loss or potential leakage of the required
transformation.
\section{Evaluation Framework}
\label{sec:eval-framework}

VLM failures on visual reasoning may arise from visual information
extraction, reasoning, or interactions between them, and these
processes are tightly coupled in IQ-style puzzles. We therefore
combine end-to-end evaluation with description-based diagnostic
analysis and two intervention-based experiments. The goal is not to
treat any single observable signal as a direct measurement of a
model's internal computation, but to examine converging evidence about
how performance changes across models, visual representations, and
inference conditions.

\subsection{End-to-End Accuracy}
\label{sec:e2e}

We measure multiple-choice accuracy across all five reasoning
categories in both domains. Because the abstract and natural domains
instantiate the same reasoning categories, these results support
category-level comparison across visual representations. The domains
are matched at the category level rather than item by item, so these
comparisons should not be interpreted as controlled contrasts between
paired instances.

\subsection{Description-Based Perception--Reasoning Attribution}
\label{sec:description-attribution-method}

For DiagVRIQ (\autoref{sec:diagnostic-annotations}), we use a
describe-then-solve protocol in which the model produces, within a
single response, (i) a natural-language description of the puzzle as
if explaining it to a non-sighted reader, followed by (ii) its
reasoning and (iii) its answer to the original multiple-choice
question (prompt in \autoref{app:describe-then-solve-prompt}).
Because the description and answer are extracted from the same
response, description correctness and answer correctness can be
analyzed on the same items.

\paragraph{Perception-success criterion.}
We compare each model-generated description with the verified
reference description using GPT-5.2~\citep{chatgpt-5.2} as a judge
(prompt in \autoref{app:description_judge_prompt}). The judge assigns
a binary \emph{perception-success} label indicating whether the
generated description correctly reports the relevant visual entities,
panel structure, and answer options present in the image. We use this
label as an operational diagnostic of the visual information expressed
in the model's description. Missing or misidentified entities,
hallucinated content, and substantive structural misunderstandings
count as failures under this criterion, regardless of whether the
final answer is correct.

\paragraph{Metrics.}
Let $P^{\checkmark}$ denote a description that satisfies the
perception-success criterion and $A^{\checkmark}$ a correct answer.
For each model and category we report:

\begin{itemize}
    \item \textbf{Description success} $P(P^{\checkmark})$: the
    fraction of items whose generated description satisfies the
    perception-success criterion.
    \item \textbf{Conditional answer accuracy}
    $P(A^{\checkmark}\mid P^{\checkmark})$: answer accuracy on items
    whose descriptions satisfy the criterion.
\end{itemize}

End-to-end accuracy decomposes exactly as
\begin{equation}
P(A^{\checkmark}) =
P(P^{\checkmark})\,P(A^{\checkmark}\mid P^{\checkmark})
+
P(P^{\times})\,P(A^{\checkmark}\mid P^{\times}).
\label{eq:decomposition}
\end{equation}

These quantities characterize the empirical relationship between
description correctness and answer correctness. We report description
success and $P(A^{\checkmark}\mid P^{\checkmark})$. A failed
description label is not a reliable signal: the description may still
have been adequate, or both the description and the answer may be
wrong. We therefore interpret the labels as diagnostic signals rather
than as a causal decomposition of perception and reasoning.

\paragraph{Judge calibration.}
We calibrate the perception-success judge against human annotations on
outputs from all five models used in the DiagVRIQ analysis:
LLaVA-1.6-Mistral-7B, Qwen2.5-VL-7B, GPT-4o, Claude Opus 5, and
GPT-5.6 Sol. Human--judge agreement is 97.1\%, 89.0\%, 92.3\%, 93.8\%,
and 91.9\%, respectively. Across all 1,045 descriptions, pooled
human--judge agreement is 92.8\%, with Cohen's $\kappa=0.86$. Because
GPT-5.2 also drafted the initial reference descriptions, we
additionally use Claude Opus 4.8 as an independent cross-family judge
on all 1,045 descriptions; the two judges agree on
97.4\% of labels.

\paragraph{Scope.}
The describe-then-solve protocol evaluates an observable generated
description rather than a model's internal visual representation. A
model may omit a detail from its description while nevertheless using
related visual information when answering the puzzle. Moreover,
although the required output format places the description before the
visible reasoning and final answer, reasoning-capable models may
perform latent computation before emitting that description.
Consequently, the description should not be interpreted as a
guaranteed causal intermediate state. We use it as an observable
diagnostic and complement it with the intervention described next.

\subsection{Ground-Truth Description Augmentation}
\label{sec:reasoning-with-gt}

To examine how performance changes when verified visual information
is made explicitly available, we re-evaluate DiagVRIQ with the
ground-truth description supplied to the model. We compare:
\begin{itemize}[leftmargin=*,itemsep=2pt,topsep=2pt]

    \item \textbf{Baseline}: accuracy from the unaugmented
    DiagVRIQ describe-then-solve condition used in
    \autoref{sec:description-attribution-method}.

    \item \textbf{GT-augmented}: the same puzzle image and
    multiple-choice question with the verified ground-truth
    description supplied as additional input and an instruction to
    use both modalities.

\end{itemize}

The ground-truth descriptions contain only observable visual
information and do not reveal the reasoning rule or correct answer
(\autoref{sec:diagnostic-annotations}; prompt in
\autoref{appendix:gt_prompt}). This intervention does not directly
modify the model's visual encoder or isolate visual perception.
Instead, it supplies verified visual information in textual form.
Any improvement may therefore reflect both access to corrected visual
content and the advantage of operating over a textual representation.
We consequently interpret the experiment as an intervention on the
information available at inference time rather than as a direct
measurement of the visual encoder.

\subsection{Tool-Augmented Inference}
\label{sec:tools-method}

Finally, we evaluate a reasoning model with and without access to a
code-execution tool that can crop, zoom, rotate, and otherwise process
images, as well as execute Python during inference
(\autoref{sec:setup}). This is a \emph{bundled} intervention:
relative to the no-tool condition, it adds image manipulation,
additional inference-time computation, and symbolic computation. We
therefore use this experiment to measure the overall benefit of
tool-augmented inference rather than to isolate the contribution of
cropping, zooming, rotation, or any individual operation.
\section{Experimental Setup}
\label{sec:setup}

\paragraph{Models.}
We evaluate 21 vision--language models (22 configurations) spanning a
wide range of model sizes, architectures, and capabilities
(full list in \autoref{app:models}):
\begin{itemize}[leftmargin=*,itemsep=2pt,topsep=2pt]

    \item \textbf{Small open-weight (3B--9B):}
    Qwen2.5-VL (3B, 3B-AWQ, 7B), InternVL3-9B,
    LLaVA-v1.6 (Mistral-7B, Vicuna-7B),
    LLaVA-OneVision-1.5-8B-RL, and Bee-8B-RL.

    \item \textbf{Mid-size open-weight (32B--34B):}
    Qwen3-VL-32B-Thinking and LLaVA-v1.6-34B.

    \item \textbf{Proprietary:}
    GPT-4o, GPT-4o-mini, GPT-5-mini, GPT-5.1, GPT-5.2,
    Gemini-2.5-Pro, and Claude Sonnet 4.6.

    \item \textbf{2026 frontier:}
    Gemini 3.1 Pro Preview, Claude Opus 5, and GPT-5.6 Sol,
    with GPT-5.6 Sol evaluated at both \emph{high} and
    \emph{medium} reasoning effort to examine sensitivity to
    inference-time reasoning.

    \item \textbf{Tool-augmented:}
    OpenAI o3 with image-manipulation tools
    (``thinking with images'')~\citep{openai-thinking-images}.

\end{itemize}

We define the \emph{2026 frontier} group as the most recent
high-capability model evaluated from each of OpenAI, Google, and
Anthropic at the time of our experiments, and refer to the remaining
17 open-weight and proprietary models as \emph{non-frontier models}.
The non-frontier group spans a broad range of model sizes,
architectures, and release dates, providing a comparison set against
which recent frontier performance can be evaluated. The open-weight
models additionally support direct reproduction under the reported
checkpoints and inference settings. GPT-5.2 is included both as an
evaluated model and as the perception-success judge; judge calibration
is reported separately in \autoref{app:calibration}.

\paragraph{Inference settings.}
Open-weight models, including Qwen3-VL-32B-Thinking, use greedy
decoding at temperature 0 on two NVIDIA RTX A6000 GPUs. Proprietary
models are accessed through their official APIs. When reasoning effort
is omitted, GPT-5.1 and GPT-5.2 use none, and o3 and GPT-5-mini use
medium. GPT-5.6 Sol is reported at high and medium effort, with medium
on DiagVRIQ. Claude Opus 5 uses adaptive thinking at effort high, and
Gemini 3.1 Pro Preview uses thinking level high. Each item is evaluated
once. Runs that sample are therefore subject to run-to-run variation.
The end-to-end prompt, image resize, and output caps are in
\autoref{app:e2e-protocol}.

\paragraph{Tool-augmented evaluation.}
We evaluate o3 through the OpenAI Responses API with the
\texttt{code\_interpreter} tool enabled, allowing the model to crop,
zoom, rotate, and otherwise process images, as well as execute Python
during inference~\citep{openai-code-interpreter}. The no-tool
condition uses the same model and task prompt without tool access.
Both conditions are evaluated once on all 1,391 VRIQ items.

\paragraph{Human baseline.}
We evaluate human performance on a stratified subset of 300 questions, with 30 questions sampled from each of the five category $\times$ two domain cells. Twenty participants (ages 18--55, mixed academic backgrounds) each solved 30 questions, and every question was independently answered by two participants. Each category--domain cell therefore contains 60 human responses. Participants had no time limit and could take breaks between items (full instructions in \autoref{app:models}). Human performance is reported as pooled response accuracy within each category--domain cell and across each domain.
\section{Results}
\label{sec:results}

We organize the results around three questions:
(i) how visual-reasoning performance varies across models and
visual domains,
(ii) how description correctness relates to answer correctness on
DiagVRIQ, and
(iii) how performance changes when models receive verified
visual information or inference-time tools.

\begin{table}[t]
\centering
\small
\caption{
Abstract-domain accuracy (\%) on VRIQ.
3D = 3D Visualization, FR = Figure Rotation,
MP = Matrix Prediction, OO = Odd One Out, and
SC = Sequence Completion.
The 2026 frontier group contains the most recent high-capability model evaluated from each of OpenAI, Google, and Anthropic at the time
of our experiments; the remaining 17 open-weight and
proprietary models are non-frontier models.
The frontier mean uses GPT-5.6 Sol (high).
Avg is the unweighted mean over the five categories.
Bold marks the best model result in each column.
Human accuracy is measured on a stratified 150-item
abstract subset.
}
\label{tab:abstract_main}
\setlength{\tabcolsep}{5pt}

\begin{tabular}{lcccccc}
\toprule
Model & 3D & FR & MP & OO & SC & Avg \\
\midrule
\multicolumn{7}{l}{\textit{Open-weight}} \\

Qwen2.5-VL-3B-Instruct
& 28.1 & 23.2 & 17.0 & 25.5 & 20.3 & 22.8 \\

Qwen2.5-VL-3B-Instruct-AWQ
& 30.0 & 17.2 & 19.1 & 19.8 & 20.3 & 21.3 \\

Qwen2.5-VL-7B-Instruct
& 32.9 & 25.3 & 31.2 & 26.4 & 40.6 & 31.3 \\

InternVL3-9B
& 31.9 & 20.2 & 31.2 & 16.0 & 24.1 & 24.7 \\

LLaVA-v1.6-Mistral-7B
& 28.6 & 20.2 & 19.9 & 27.4 & 15.8 & 22.4 \\

LLaVA-v1.6-Vicuna-7B
& 23.8 & 16.2 & 18.4 & 15.1 & 23.3 & 19.4 \\

LLaVA-OneVision-1.5-8B-RL
& 31.9 & 22.2 & 41.8 & 24.5 & 14.3 & 27.0 \\

Bee-8B-RL
& 21.9 & 29.8 & 41.8 & 24.5 & 30.8 & 29.8 \\

Qwen3-VL-32B-Thinking
& 33.8 & 31.3 & 34.0 & 34.0 & 44.4 & 35.5 \\

LLaVA-v1.6-34B
& 22.9 & 20.2 & 26.2 & 19.8 & 24.8 & 22.8 \\

\midrule
\multicolumn{7}{l}{\textit{Proprietary}} \\

GPT-4o
& 31.0 & 31.8 & 29.8 & 30.2 & 28.6 & 30.3 \\

GPT-4o-mini
& 24.8 & 25.3 & 23.4 & 28.3 & 23.3 & 25.0 \\

GPT-5-mini
& 21.9 & 24.2 & 25.5 & 26.4 & 24.8 & 24.6 \\

GPT-5.1
& 27.1 & 30.8 & 27.0 & 34.0 & 27.8 & 29.3 \\

Gemini-2.5-Pro
& 29.5 & 31.3 & 27.7 & 27.4 & 27.1 & 28.6 \\

Claude Sonnet 4.6
& 21.9 & 29.8 & 32.6 & 22.6 & 24.1 & 26.2 \\

GPT-5.2
& 36.2 & 39.9 & 44.0 & 32.1 & 25.6 & 35.5 \\

\midrule
\multicolumn{7}{l}{\textit{2026 frontier}} \\

Gemini 3.1 Pro Preview
& 35.2 & 55.1 & 60.3 & 30.2 & 25.6 & 41.3 \\

Claude Opus 5
& 23.8 & 76.8 & 46.1 & 60.4 & 27.8 & 47.0 \\

GPT-5.6 Sol (medium)
& \textbf{50.0}
& \textbf{84.8}
& \textbf{85.1}
& 51.9
& 64.7
& 67.3 \\

GPT-5.6 Sol (high)
& 44.8
& \textbf{84.8}
& 80.1
& \textbf{65.1}
& \textbf{79.7}
& \textbf{70.9} \\

\midrule

Mean, non-frontier (17)
& 28.1 & 25.8 & 28.9 & 25.5 & 25.9 & 26.8 \\

Mean, 2026 frontier (3)
& 34.6 & 72.2 & 62.2 & 51.9 & 44.4 & 53.1 \\

Human baseline
& 85.0 & 93.3 & 88.3 & 91.7 & 93.3 & 90.3 \\

\bottomrule
\end{tabular}
\end{table}
\begin{table}[t]
\centering
\small
\caption{
Natural-domain accuracy (\%) on VRIQ.
Category abbreviations and model groups follow
\autoref{tab:abstract_main}.
The frontier mean uses GPT-5.6 Sol (high).
Avg is the unweighted mean over the five categories.
Bold marks the best model result in each column.
Human accuracy is measured on a stratified 150-item
natural subset.
}
\label{tab:natural_main}
\setlength{\tabcolsep}{5pt}

\begin{tabular}{lcccccc}
\toprule
Model & 3D & FR & MP & OO & SC & Avg \\
\midrule
\multicolumn{7}{l}{\textit{Open-weight}} \\

Qwen2.5-VL-3B-Instruct
& 29.7 & 40.2 & 43.2 & 40.9 & 34.0 & 37.6 \\

Qwen2.5-VL-3B-Instruct-AWQ
& 25.2 & 35.1 & 28.8 & 36.2 & 37.0 & 32.5 \\

Qwen2.5-VL-7B-Instruct
& 29.7 & 35.1 & 39.7 & 49.7 & 25.0 & 35.8 \\

InternVL3-9B
& 39.6 & 45.4 & 34.9 & 40.3 & 40.0 & 40.0 \\

LLaVA-v1.6-Mistral-7B
& 27.0 & 17.5 & 24.7 & 30.2 & 20.0 & 23.9 \\

LLaVA-v1.6-Vicuna-7B
& 18.9 & 15.5 & 15.1 & 34.2 & 24.0 & 21.5 \\

LLaVA-OneVision-1.5-8B-RL
& 36.9 & 45.4 & 52.1 & 28.9 & 41.0 & 40.8 \\

Bee-8B-RL
& 50.5 & 29.9 & 45.9 & 24.8 & 33.0 & 36.8 \\

Qwen3-VL-32B-Thinking
& 41.4 & 49.5 & 50.0 & 53.7 & 42.0 & 47.3 \\

LLaVA-v1.6-34B
& 27.0 & 19.6 & 43.2 & 44.3 & 40.0 & 34.8 \\

\midrule
\multicolumn{7}{l}{\textit{Proprietary}} \\

GPT-4o
& 47.7 & 40.2 & 62.3 & 57.0 & 64.0 & 54.3 \\

GPT-4o-mini
& 38.7 & 40.2 & 47.9 & 48.3 & 44.0 & 43.8 \\

GPT-5-mini
& 27.9 & 40.2 & 50.7 & 48.3 & 49.0 & 43.2 \\

GPT-5.1
& 41.4 & 40.2 & 58.9 & 59.7 & 59.0 & 51.9 \\

Gemini-2.5-Pro
& 58.6 & 33.0 & 65.1 & 49.7 & 61.0 & 53.5 \\

Claude Sonnet 4.6
& 44.1 & 25.8 & 63.7 & 55.7 & 72.0 & 52.3 \\

GPT-5.2
& 39.6 & 46.4 & 56.2 & 51.7 & 68.0 & 52.4 \\

\midrule
\multicolumn{7}{l}{\textit{2026 frontier}} \\

Gemini 3.1 Pro Preview
& 55.0
& \textbf{80.4}
& \textbf{89.7}
& \textbf{96.0}
& 70.0
& \textbf{78.2} \\

Claude Opus 5
& 50.5 & 74.2 & 86.3 & 89.9 & 82.0 & 76.6 \\

GPT-5.6 Sol (medium)
& \textbf{60.4}
& 35.1
& 80.1
& 79.9
& 75.0
& 66.1 \\

GPT-5.6 Sol (high)
& 55.0
& 35.1
& 80.1
& 79.9
& \textbf{85.0}
& 67.0 \\

\midrule

Mean, non-frontier (17)
& 36.7 & 35.2 & 46.0 & 44.3 & 44.3 & 41.3 \\

Mean, 2026 frontier (3)
& 53.5 & 63.2 & 85.4 & 88.6 & 79.0 & 73.9 \\

Human baseline
& 96.7 & 98.3 & 96.7 & 98.3 & 98.3 & 97.7 \\

\bottomrule
\end{tabular}
\end{table}

\subsection{End-to-End Performance}
\label{sec:e2e-results}

\paragraph{Non-frontier models remain near chance on abstract puzzles.}
Averages below are unweighted means of the five category accuracies.\footnote{Per-cell 95\% Wilson confidence intervals are reported in \autoref{app:e2e-ci}; their half-widths range from about $\pm$3 to $\pm$10 points depending on accuracy and category size.}
Across the 17 non-frontier models, that mean is
26.8\% on abstract puzzles
(\autoref{tab:abstract_main}), with the strongest models reaching
35.5\%. Category-level averages range from 25.5\% to 28.9\%,
indicating that no abstract category is consistently easy.
Proprietary models also do not uniformly outperform open-weight
models: GPT-4o reaches 30.3\%, compared with 35.5\% for
Qwen3-VL-32B-Thinking. Human performance remains substantially
higher.

\paragraph{The 2026 frontier improves substantially, but unevenly.}
On abstract puzzles, the unweighted category means are 70.9\% for
GPT-5.6 Sol at high reasoning effort, 47.0\% for Claude Opus 5, and
41.3\% for Gemini 3.1 Pro Preview. Their category profiles also differ sharply. Figure
Rotation rises from a 25.8\% average among non-frontier models to
55.1--84.8\% for the frontier models, whereas 3D Visualization
remains difficult, with a frontier average of 34.6\%. Gemini 3.1 Pro
Preview remains near the non-frontier mean on abstract Sequence
Completion (25.6\%) and Odd One Out (30.2\%).

\paragraph{The benefit of natural images is model-dependent.}
Across the non-frontier models, average accuracy increases from 26.8\%
on abstract puzzles to 41.3\% on natural puzzles
(\autoref{tab:natural_main}), with the largest category-level gains in
Matrix Prediction ($+17.1$ points), Odd One Out ($+18.8$), and
Sequence Completion ($+18.4$). The cross-domain difference is larger
for Claude Opus 5 ($+29.6$) and Gemini 3.1 Pro Preview ($+36.9$).
GPT-5.6 Sol shows a different pattern, reaching 67.0\% on natural
puzzles versus 70.9\% on abstract puzzles. This difference is partly
associated with natural Figure Rotation, where Sol reaches 35.1\%,
compared with 74.2--80.4\% for the other frontier models. Natural
images therefore do not provide a uniform advantage, and all models
remain below human performance.

\paragraph{Changing reasoning effort yields modest, mixed differences.}
Increasing GPT-5.6 Sol from medium to high reasoning effort changes
average accuracy from 67.3\% to 70.9\% on abstract puzzles and from
66.1\% to 67.0\% on natural puzzles. The largest positive changes
occur on abstract Sequence Completion ($+15.0$ points), abstract Odd
One Out ($+13.2$), and natural Sequence Completion ($+10.0$).
By contrast, 3D Visualization decreases by approximately 5 points in
both domains, as does abstract Matrix Prediction ($-5.0$), while
Figure Rotation in both domains and natural Matrix Prediction and Odd
One Out are unchanged in this run. Each setting is a single sample
without a confidence interval, so we treat the comparison as a
sensitivity check.

\subsection{Description-Based Analysis on DiagVRIQ}
\label{sec:description-attribution}

\begin{table}[t]
\centering
\small
\caption{Pooled DiagVRIQ metrics over all 209 items, with 95\% Wilson
intervals. Description success is the fraction of generated descriptions
satisfying the perception-success criterion. Acc.\ $\mid
P^{\checkmark}$ denotes answer accuracy conditional on the generated
description satisfying that criterion.}
\label{tab:diag_pooled}
\setlength{\tabcolsep}{5pt}

\begin{tabular}{lccc}
\toprule
Model & End-to-end & \makecell{Description\\success}
& Acc.\ $\mid P^{\checkmark}$ \\
\midrule

LLaVA-v1.6-Mistral-7B
& 26.8 [21.2, 33.2]
& 12.0 [8.2, 17.1]
& 52.0 [33.5, 70.0] \\

Qwen2.5-VL-7B
& 51.7 [44.9, 58.4]
& 62.2 [55.5, 68.5]
& 69.2 [60.8, 76.5] \\

GPT-4o
& 84.7 [79.2, 88.9]
& 92.3 [87.9, 95.2]
& 90.7 [85.7, 94.0] \\

Claude Opus 5
& 90.9 [86.2, 94.1]
& 85.6 [80.2, 89.8]
& 91.1 [86.0, 94.4] \\

GPT-5.6 Sol
& 88.0 [82.9, 91.8]
& 84.7 [79.2, 88.9]
& 88.1 [82.5, 92.1] \\

\bottomrule
\end{tabular}
\end{table}

\autoref{tab:diag_pooled} summarizes performance over all 209
DiagVRIQ items with 95\% confidence intervals; per-category results
are reported in \autoref{tab:perc_reas_acc_updated}. GPT-5.6 Sol is
evaluated at medium reasoning effort.

\paragraph{Description success varies substantially across models.}
The fraction of generated descriptions satisfying the
perception-success criterion ranges from 12.0\% for LLaVA-1.6 and
62.2\% for Qwen2.5-VL-7B to 92.3\% for GPT-4o, with Claude Opus 5
and GPT-5.6 Sol reaching 85.6\% and 84.7\%, respectively.
Category-level results reveal additional variation. LLaVA performs
particularly poorly on Matrix Prediction (6.1\%) and Sequence
Completion (3.3\%), while Qwen reaches 70.4\% on Matrix Prediction
and 84.3\% on Odd One Out but only 30.0\% on Sequence Completion,
which is the lowest-description-success category for four of the five
models. Description success therefore increases sharply from the
weakest to the stronger systems, although the ordering is not
monotonic among GPT-4o, Claude Opus 5, and GPT-5.6 Sol.

\paragraph{Correct descriptions are not sufficient for correct answers.}
Among items whose descriptions satisfy the criterion, LLaVA-1.6
answers 52.0\% correctly and Qwen2.5-VL-7B 69.2\%, compared with
90.7\% for GPT-4o, 91.1\% for Claude Opus 5, and 88.1\% for
GPT-5.6 Sol. Thus, even among items with correct generated
descriptions, the strongest models still answer approximately
9--12\% incorrectly. This suggests that accurate visual description
alone is not sufficient for reliable task performance.

\subsection{Ground-Truth Description Augmentation}
\label{sec:gt-results}

\begin{table}[t]
\centering
\small
\caption{DiagVRIQ accuracy (\%) without (Base) and with (GT) ground-truth description augmentation, and the change $\Delta$ (points). Every model--category cell improves, although the GPT-4o gains are small.}
\label{tab:gt-augmentation}
\setlength{\tabcolsep}{4.5pt}
\begin{tabular}{lccccccccc}
\toprule
& \multicolumn{3}{c}{Matrix Prediction} & \multicolumn{3}{c}{Odd One Out} & \multicolumn{3}{c}{Sequence Completion} \\
\cmidrule(lr){2-4}\cmidrule(lr){5-7}\cmidrule(lr){8-10}
Model & Base & GT & $\Delta$ & Base & GT & $\Delta$ & Base & GT & $\Delta$ \\
\midrule
LLaVA-v1.6-Mistral-7B & 21.4 & 36.7 & +15.3 & 39.2 & 49.0 & +9.8 & 25.0 & 66.7 & +41.7 \\
Qwen2.5-VL-7B         & 59.2 & 69.4 & +10.2 & 52.9 & 56.9 & +4.0 & 38.3 & 66.7 & +28.4 \\
GPT-4o                & 87.8 & 89.8 & +2.0  & 84.3 & 90.2 & +5.9 & 80.0 & 83.3 & +3.3 \\
\bottomrule
\end{tabular}
\end{table}

Providing a verified, rule-neutral description improves accuracy in
every reported model--category cell
(\autoref{tab:gt-augmentation}), although the GPT-4o gains are small. Claude Opus 5 and GPT-5.6 Sol are
not part of this intervention. LLaVA gains $+15.3$ points on
Matrix Prediction, $+9.8$ on Odd One Out, and $+41.7$ on Sequence
Completion; Qwen improves by $+10.2$, $+4.0$, and $+28.4$ points,
respectively; and GPT-4o, whose generated descriptions are already
highly accurate, improves by only 2.0--5.9 points. Errors nevertheless
remain: GPT-4o still misses approximately 10--17\% of items across the
three categories, while Qwen remains below 57\% on Odd One Out.

\subsection{Tool-Augmented Inference}
\label{sec:tool-results}

\begin{table}[t]
\centering
\small
\caption{Accuracy (\%) of o3 with and without tool access on all
VRIQ items. Natural-domain category sizes are 111 (3D), 97 (FR),
146 (MP), 149 (OO), and 100 (SC); abstract-domain category sizes are
210, 198, 141, 106, and 133, respectively. Overall is the unweighted
mean over the five categories.}
\label{tab:o3-tools}
\setlength{\tabcolsep}{6pt}
\begin{tabular}{lcccccc}
\toprule
& \multicolumn{3}{c}{Natural domain} & \multicolumn{3}{c}{Abstract domain} \\
\cmidrule(lr){2-4}\cmidrule(lr){5-7}
Category & With & Without & $\Delta$ & With & Without & $\Delta$ \\
\midrule
3D Visualization    & 96.4  & 48.6 & +47.8 & 35.2 & 23.8 & +11.4 \\
Figure Rotation     & 91.8  & 40.2 & +51.6 & 68.2 & 41.9 & +26.3 \\
Matrix Prediction   & 75.3  & 62.3 & +13.0 & 46.8 & 35.5 & +11.3 \\
Odd One Out         & 69.8  & 59.1 & +10.7 & 52.8 & 31.1 & +21.7 \\
Sequence Completion & 80.0  & 69.0 & +11.0 & 48.9 & 36.8 & +12.1 \\
\midrule
Overall             & 82.7  & 55.8 & +26.9 & 50.4 & 33.8 & +16.6 \\
\bottomrule
\end{tabular}
\end{table}

Tool access substantially improves o3 performance: natural-domain
accuracy increases from 55.8\% to 82.7\% ($+26.9$ points), while
abstract-domain accuracy increases from 33.8\% to 50.4\%
($+16.6$; \autoref{tab:o3-tools}). The largest gains occur on
spatially demanding tasks: natural Figure Rotation ($+51.6$),
natural 3D Visualization ($+47.8$), and abstract Figure Rotation
($+26.3$). Matrix Prediction and Sequence Completion improve by
approximately $+11$ to $+13$ points across the two domains.
Representative traces in \autoref{app:o3-traces} illustrate both
successful uses of image manipulation and persistent counting,
localization, and rule-identification errors.
\section{Discussion and Conclusion}
\label{sec:discussion}

\paragraph{Summary of findings.}
Our results show that visual-reasoning performance has improved
substantially at the frontier, but the gains are highly uneven across
models, domains, and reasoning categories. The 17 non-frontier models
remain near chance on abstract puzzles (26.8\% unweighted category mean),
whereas GPT-5.6 Sol reaches 70.9\% on the same average. This improvement is not shared uniformly by
the other 2026 frontier models, and all evaluated models remain below
human performance. The effect of visual representation is also
model-dependent: natural-image performance is substantially higher
than abstract-domain performance for Claude Opus 5 and Gemini 3.1 Pro
Preview, while GPT-5.6 Sol performs similarly across the two domains.

DiagVRIQ further shows that end-to-end accuracy can conceal different
failure patterns. Description success varies substantially across
models, rising from 12.0\% for LLaVA-1.6 and 62.2\% for
Qwen2.5-VL-7B to above 84\% for the stronger systems. At the same
time, a correct description does not guarantee a correct final answer:
even among the strongest models, approximately 9--12\% of items with
correct descriptions are still answered incorrectly. These results
show that improvements in visual description quality do not eliminate
errors in downstream visual reasoning.

The intervention experiments provide complementary evidence.
Supplying verified, rule-neutral visual descriptions improves accuracy
in every reported cell (LLaVA-1.6, Qwen2.5-VL-7B, and GPT-4o), with
the largest gains for LLaVA and Qwen. However, errors remain even when verified visual
information is supplied. Tool-augmented inference also produces large
gains, particularly on spatially demanding tasks, although this setting
combines image manipulation with additional computation and code
execution. Finally, increasing GPT-5.6 Sol from medium to high
reasoning effort produces only modest average changes, indicating that
greater inference-time reasoning does not uniformly improve
visual-reasoning performance.

\paragraph{Conclusion.}
We introduced VRIQ, a benchmark of 1{,}391 IQ-style visual reasoning
puzzles spanning category-matched abstract and natural-image domains,
together with DiagVRIQ, a 209-item diagnostic subset with verified,
rule-neutral textual descriptions. Across 17 non-frontier models and
three 2026 frontier systems, visual-reasoning performance improves
substantially at the frontier but remains strongly dependent on the
model, visual representation, and reasoning category. The
description-based analysis further shows that accurate visual
description alone is not sufficient for reliable task performance,
while the intervention experiments demonstrate that both verified
visual information and additional inference-time tools can improve
accuracy without eliminating errors. VRIQ therefore complements
end-to-end benchmarking with diagnostic and intervention-based
analyses of visual-reasoning behavior.

\section*{Limitations}

\paragraph{Scope of visual reasoning.}
VRIQ evaluates five categories of IQ-style visual reasoning, including
relational comparison, spatial transformation, pattern completion, and
visual discrimination. These controlled tasks are useful for systematic
comparison across models and visual domains, but they do not cover the
full range of real-world multimodal reasoning, including temporal,
causal, interactive, or embodied reasoning.

\paragraph{Potential data contamination.}
Some abstract puzzles are adapted from publicly accessible examination
materials that may have appeared in model training data. To reduce
this risk, adapted items were modified in shape, count, orientation, or
styling while keeping the rule, and other abstract items, including the
Python-generated Figure Rotation questions, were written by the
authors. Natural-domain questions were written for this benchmark,
and their images were either generated or taken from public sources.
Nevertheless, residual
contamination cannot be ruled out completely, particularly for models
whose training data are not publicly documented.

\paragraph{Description-based diagnostic scope.}
The describe-then-solve protocol evaluates the correctness of an
observable model-generated description rather than providing direct
access to a model's internal visual representation. In addition,
although the required output format places the description before the
visible reasoning and final answer, reasoning-capable models may
perform latent computation before producing that description.
Accordingly, the description should be interpreted as a diagnostic
output rather than as a guaranteed causal intermediate state between
visual input and final answer.

\paragraph{Intervention effects.}
The intervention experiments evaluate combined inference settings
rather than isolated mechanisms. Ground-truth description augmentation
provides verified visual information in textual form, so improvements
may reflect both the supplied visual information and the use of a
textual representation. Tool-augmented inference combines image
manipulation with additional computation and code execution. The
reported gains therefore characterize the overall intervention
settings rather than the contribution of any individual component.

\subsubsection*{Broader Impact Statement}

VRIQ and DiagVRIQ are intended for research on the evaluation and
diagnosis of visual reasoning in vision--language models. Benchmark
images are either authored by the research team or obtained from
publicly available, copyright-compliant sources, and natural-domain
images were screened to exclude identifiable individuals, personal
information, and offensive content. The benchmark and annotations
are available at \linkVRIQ\ and \linkDiag\
under CC BY-NC 4.0 for non-commercial research use,
and the evaluation code is available at \linkCode,
and we discourage their use for model training in order to preserve
their value as evaluation data.

The human-baseline study involved adult volunteers who were informed
of the task and the use of their responses; no personally identifying
information was collected or retained. Our use of AI assistants in
item proposal, description drafting, and evaluation is disclosed in
\autoref{app:models}, and final benchmark items and reported analyses
were reviewed or verified by the human researchers.

\subsubsection*{Acknowledgments}

We thank Ara Esfarjani, Qingyi Jiang, and Brandon Cui for their help with item authoring, image sourcing, layout assembly, and
answer-correctness verification.

\bibliography{main}
\bibliographystyle{tmlr}

\appendix
\section{Procedure for Natural Question Generation}
\label{app:generation-procedure}

Natural-domain questions were written by the authors. Images were either generated with GPT-5.1 or Gemini 2.5 Flash, or drawn from publicly available, copyright-compliant sources. Human annotators selected, edited, and verified every retained item. Because GPT-5.1 is also an evaluated model and Gemini 2.5 Flash belongs to a family we evaluate, we do not treat those models' natural-domain scores as independent of the image source.

\paragraph{Worked example.}
For Sequence Completion, one generated biology-themed instance used
the sequence \emph{frog egg, tadpole, tadpole with legs} with
candidate options \emph{frog egg, adult frog, fish, lizard}. The
intended rule was continuation of the frog life cycle, making
\emph{adult frog} the correct answer. Annotators obtained an image
for each entity and assembled the final puzzle.

\paragraph{Annotator team and compensation.}
Throughout this paper, ``annotators'' refers to the team that contributed to item authoring, image sourcing, layout assembly, and answer-correctness verification across both the natural and abstract domains (\autoref{app:generation-procedure}, \autoref{app:gt-description-protocol}, and \autoref{app:abstract-sourcing}). The team comprised the paper's authors together with three undergraduate research assistants who provided support across these activities. The undergraduate assistants were compensated as paid student researchers at the standard hourly rate set by our institution.

\section{Ground-Truth Description Drafting and Verification Protocol}
\label{app:gt-description-protocol}

DiagVRIQ ground-truth descriptions were generated by GPT-5.2~\citep{chatgpt-5.2} under three constraints embedded in the generation prompt: (i) describe all panels, entities, attributes, and answer options visible in the image; (ii) do not name the underlying transformation, grouping, or sequence rule; and (iii) do not reveal the correct answer or its option label.

Each generated description was then independently verified by an annotator against the puzzle image. The annotator checked \textbf{coverage} (all panels and answer options are described), \textbf{rule-neutrality} (no statement of the puzzle's rule, transformation, or grouping principle), and \textbf{answer-neutrality} (no privileged reference to any specific option). Descriptions that failed any check were revised before inclusion in DiagVRIQ.

For Sequence Completion, descriptions enumerate the contents of each top-row panel and each candidate answer panel using concrete visual vocabulary (shape, position, orientation, count) without indicating how panels relate. For Matrix Prediction, descriptions enumerate each cell and answer option. For Odd One Out, descriptions list each candidate entity together with its visible attributes, without identifying which property the puzzle groups by.

\section{Abstract Puzzle Sourcing and Modification}
\label{app:abstract-sourcing}

\subsection{Source Materials and Screening}

Abstract items come from public aptitude and civil-service materials, reformatted and modified in shape, count, orientation, or styling while keeping the rule, and from puzzles written by the authors. The 3D items were built in Tinkercad~\citep{tinkercad}. Abstract Figure Rotation questions were generated by Python code.

Annotators manually filtered sourced items to remove puzzles that (i) admitted ambiguous or non-unique answers, (ii) were visually low quality or poorly legible, or (iii) did not match the target reasoning categories. All retained puzzles were independently reviewed by a second annotator.

\subsection{Reformatting and Modification}

Sourced puzzles were reconstructed into a unified VRIQ format. Panels and answer options were normalized into consistent configurations, and minor visual artifacts were removed when necessary.

To reduce contamination risk from widely circulated puzzle collections, sourced items were modified into new instances by altering shape identities, counts, orientations, or styling while preserving the underlying reasoning rule.

\subsection{Manually Authored Items}

In addition to sourced materials, the authors wrote abstract puzzles from scratch, including the Figure Rotation questions generated by Python code. The 3D items were built in Tinkercad~\citep{tinkercad}. These items followed the same quality-control pipeline as sourced puzzles.

\paragraph{License and intended use.}
VRIQ, DiagVRIQ, and the ground-truth descriptions
are available at \linkVRIQ\ and \linkDiag\
under the Creative Commons Attribution-NonCommercial 4.0 International (CC BY-NC 4.0) license.
Evaluation code
is available at \linkCode\
under the MIT license. The benchmark is intended for research use in evaluating and analyzing visual reasoning capabilities of vision--language models. We discourage use as a training set, as this would compromise the benchmark's diagnostic value for future model evaluation.

\begin{figure*}[t]
    \centering
    \includegraphics[width=\textwidth]{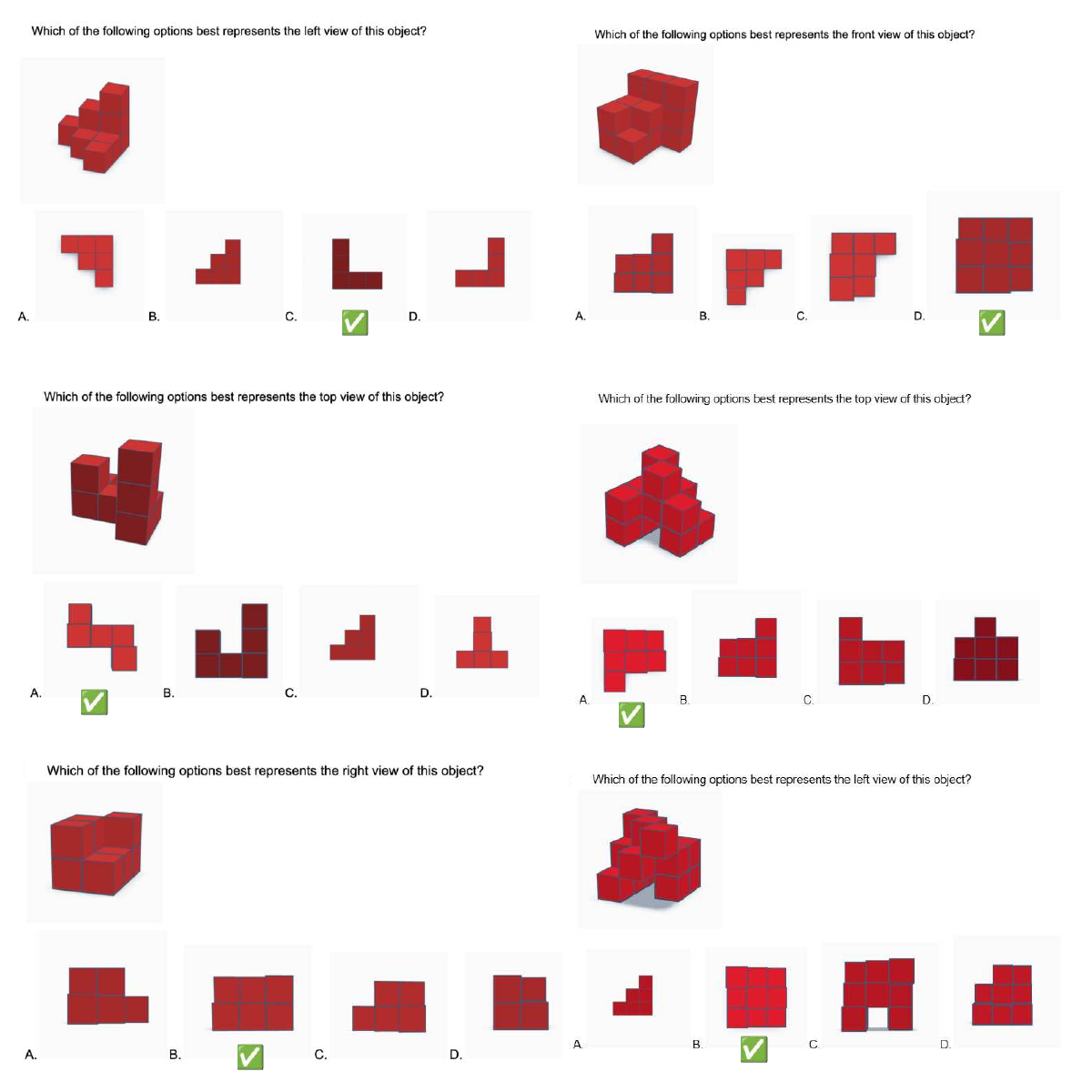}
    \caption{3D abstract examples built by Tinkercad.}
    \label{fig:3d-abstract-samples}
\end{figure*}

\section{Models and Evaluation Details}
\label{app:models}

\autoref{tab:models} lists all evaluated VLMs with parameter counts, model types, and source references.

\begin{table*}[h]
\centering
\small
\caption{Vision–language models evaluated on VRIQ. Open-weight models accessed via Hugging Face; proprietary models accessed via official APIs. ``Thinking'' indicates reasoning-optimized variants; ``Tools'' indicates models with image-manipulation tool access during inference.}
\label{tab:models}
\setlength{\tabcolsep}{6pt}
\begin{tabular}{lllc}
\toprule
Model & Params & Type & Citation \\
\midrule
\multicolumn{4}{l}{\textit{Small open-weight (3B--9B)}} \\
Qwen2.5-VL-3B-Instruct          & 3B   & Instruction-tuned & \citep{Qwen2.5-VL-3B-Instruct} \\
Qwen2.5-VL-3B-Instruct-AWQ      & 3B   & Quantized         & \citep{Qwen2.5-VL-3B-Instruct-AWQ} \\
Qwen2.5-VL-7B-Instruct          & 7B   & Instruction-tuned & \citep{Qwen2.5-VL-7B-Instruct} \\
InternVL3-9B                    & 9B   & Instruction-tuned & \citep{InternVL3-9B} \\
LLaVA-v1.6-Mistral-7B           & 7B   & Instruction-tuned & \citep{llava-v1.6-mistral-7b-hf} \\
LLaVA-v1.6-Vicuna-7B            & 7B   & Instruction-tuned & \citep{llava-v1.6-vicuna-7b-hf} \\
LLaVA-OneVision-1.5-8B-RL       & 8B   & RL-tuned          & \citep{an2025llava} \\
Bee-8B-RL                       & 8B   & RL-tuned          & \citep{zhang2025bee} \\
\midrule
\multicolumn{4}{l}{\textit{Mid-size open-weight (32B--34B)}} \\
Qwen3-VL-32B-Thinking           & 32B  & Thinking          & \citep{bai2025qwen3vltechnicalreport} \\
LLaVA-v1.6-34B                  & 34B  & Instruction-tuned & \citep{llava-v1.6-34b-hf} \\
\midrule
\multicolumn{4}{l}{\textit{Non-frontier proprietary}} \\
GPT-4o                          & --   & Instruction-tuned & \citep{chatgpt-4o} \\
GPT-4o-mini                     & --   & Instruction-tuned & \citep{chatgpt-4o-mini} \\
GPT-5.1                         & --   & Proprietary          & \citep{chatgpt-5.1} \\
GPT-5-mini                      & --   & Proprietary          & \citep{chatgpt-5-mini} \\
Gemini-2.5-Pro                  & --   & Proprietary          & \citep{comanici2025gemini} \\
Claude Sonnet 4.6               & --   & Proprietary          & \citep{anthropic2026claude46} \\
GPT-5.2                         & --   & Proprietary          & \citep{chatgpt-5.2} \\
\midrule
\multicolumn{4}{l}{\textit{Proprietary, 2026 frontier}} \\
Gemini 3.1 Pro Preview          & --   & Thinking          & \citep{gemini-3.1-pro} \\
Claude Opus 5                   & --   & Thinking          & \citep{anthropic2026opus5} \\
GPT-5.6 Sol (high / medium effort) & -- & Thinking         & \citep{openai2026gpt56} \\
\midrule
\multicolumn{4}{l}{\textit{Tool-augmented}} \\
OpenAI o3 (thinking with images) & --  & Thinking + Tools  & \citep{openai-o3} \\
\bottomrule
\end{tabular}
\end{table*}

\paragraph{Compute budget.}
Open-source models were evaluated on a single workstation with 2 NVIDIA RTX A6000 GPUs (48GB VRAM each). Proprietary models (GPT-4o, GPT-4o-mini, GPT-5-mini, GPT-5.1, GPT-5.2, Gemini-2.5-Pro, Gemini 3.1 Pro Preview, Claude Sonnet 4.6, Claude Opus 5, GPT-5.6 Sol, and o3) were accessed through their official APIs. Open-weight models, including Qwen3-VL-32B-Thinking, use greedy decoding at temperature 0. When reasoning effort is omitted, GPT-5.1 and GPT-5.2 use none, and models released before GPT-5.1, including o3 and GPT-5-mini, use medium. Each item is evaluated once unless stated otherwise, and no hyperparameter search was performed. Runs that sample are subject to run-to-run variation.

\subsection{End-to-End Prompt and Image Handling}
\label{app:e2e-protocol}

The main accuracy tables use the puzzle image together with the item's question. The model is instructed to write its reasoning inside \texttt{<think>} tags and the final option inside \texttt{<answer>} tags. Images are resized so that the total number of pixels is at most 250{,}000 and each side is at least 28 pixels. GPT-5.6 Sol uses an output cap of 16{,}384 tokens. The shared evaluation script caps other API generations at 1{,}024 tokens, and the logged open-weight runs use 512 tokens.

\paragraph{Human baseline details.}
The 300-question subset was selected by stratified sampling (30 items from each of 5 categories $\times$ 2 domains). Twenty adult participants (ages 18--55, mixed academic backgrounds) each solved 30 questions, so every question was independently answered by two participants.

\paragraph{Participant instructions.}
Participants were given the following written instructions before beginning the task: ``You will see a series of visual reasoning IQ puzzles, each with four answer options. For each puzzle, select the option that best completes the pattern. There is no time limit, and you may take breaks between items. If you are uncertain, select your best guess. Your responses contribute to an aggregate human-accuracy estimate; no personally identifying information is recorded.'' Participants completed the task in a single session without per-item feedback. No training items or calibration questions were provided.

\paragraph{Recruitment, payment, and consent.}
Participants were unpaid volunteers and were informed that participation was entirely voluntary. Before beginning the task, they were informed that their responses would be used solely to compute aggregate human-baseline accuracy and that no personally identifying information would be recorded or retained.

\paragraph{Ethics review.}
The human-baseline data collection involved adult volunteers solving multiple-choice puzzles with no personally identifying information collected or retained, and was determined not to require formal IRB review at our institution.

\paragraph{Use of AI assistants.}
GPT-5.2 was used to generate initial DiagVRIQ ground-truth descriptions, which were subsequently verified and revised by human annotators (\autoref{app:gt-description-protocol}); GPT-5.2 was used as the LLM-as-judge for perception-success labeling; and Claude Opus 4.8 was used as an independent cross-family judge to validate those labels. Calibration against human annotations is reported in \autoref{app:calibration}. AI assistants were also used for minor stylistic editing of the manuscript. All technical claims, analyses, and final benchmark items were authored or verified by the human researchers.

\section{Describe-Then-Solve Prompt}
\label{app:describe-then-solve-prompt}

In the describe-then-solve protocol (\autoref{sec:description-attribution-method}), the model produces a description of the puzzle followed by an answer, in a single generation. We use the prompt in \autoref{fig:describe_then_solve_prompt} for all evaluated VLMs. The description and answer are extracted from the same generation via tagged delimiters; no second model call is required.

\begin{tcolorbox}[
    enhanced,
    breakable,
    colback=gray!5,
    colframe=gray!50,
    boxrule=0.5pt
]
\small

\textbf{System.}
You are an expert at solving visual reasoning puzzles. You will be shown an image and a multiple-choice question.

Before answering, produce a faithful description of the image as if explaining it to a non-sighted reader. Describe all panels, entities, attributes, and answer options. Do not infer the rule or answer during the description.

Then reason about the puzzle and provide the final answer.

\medskip

\textbf{User.}

\texttt{[image]}

\textbf{Question:} \texttt{\{question\}}

\textbf{Options:}
A) \texttt{\{a\}}
B) \texttt{\{b\}}
C) \texttt{\{c\}}
D) \texttt{\{d\}}

\medskip

Format:

\texttt{<description>...</description>}\\
\texttt{<think>...</think>}\\
\texttt{<answer>X</answer>}

where \texttt{X} is one of A, B, C, or D.

\end{tcolorbox}

\captionof{figure}{Describe-then-solve prompt used for perception--reasoning attribution (\autoref{sec:description-attribution-method}). The model produces the description, reasoning, and answer in a single generation.}
\label{fig:describe_then_solve_prompt}

\section{Description-Quality Judge Prompt}
\label{app:description_judge_prompt}

We use GPT-5.2 as the LLM judge. For each DiagVRIQ item, the judge receives (i) the puzzle image, (ii) the verified ground-truth description (\autoref{app:gt-description-protocol}), and (iii) the model-generated description from the describe-then-solve protocol. It returns a brief justification and a binary perception-success label, which is the attribution signal used throughout the paper. The same prompt was used for the cross-family validation judge. The full prompt is shown below.

\begin{tcolorbox}[
    enhanced,
    breakable,
    colback=gray!5,
    colframe=gray!50,
    boxrule=0.5pt,
    arc=1.5mm
]
\small

\textbf{System.}
You are an expert evaluator of vision--language model descriptions for visual reasoning tasks. You will be given a puzzle image, a human-verified reference description, and a model-generated description. Your task is to determine whether the model description correctly captures the visual content of the image.

\medskip

\textbf{Important.}
Do \emph{not} evaluate whether the description contains enough information to solve the puzzle or infer the underlying rule. Evaluate only whether it accurately reports what is visually present.

\medskip

\textbf{User.}

\texttt{[image]}

\medskip

\textbf{Reference description:}
\texttt{\{ground\_truth\}}

\medskip

\textbf{Model description:}
\texttt{\{model\_description\}}

\medskip

\textbf{Perception Success (binary).}

Does the model description correctly identify the visual entities, panel structure, and answer options shown in the image?

\begin{itemize}
\item \texttt{"yes"}: entities, panels, and answer options are correctly identified; minor wording differences or small omissions are acceptable.
\item \texttt{"no"}: entities are misidentified, hallucinated, omitted, or the panel structure is misunderstood.
\end{itemize}

Provide a brief justification (1--2 sentences), then return the verdict.

\medskip

Return \textbf{only} valid JSON:

\begin{verbatim}
{
  "reason": "...",
  "perception_success": "yes/no"
}
\end{verbatim}

\end{tcolorbox}

Following Prometheus-Vision~\citep{lee2024prometheus}, the prompt
states explicit criteria for each verdict and asks for a short
justification before the verdict. The criterion evaluates description
correctness: substantive omissions, misidentifications, hallucinations,
or structural misunderstandings count as failures, while minor wording
differences and small omissions are allowed. The resulting label is an
observable measure of the correctness of the generated description and
should not be interpreted as direct access to the model's internal
visual representation.

\section{Calibration of the Perception-Success Judge}
\label{app:calibration}

To validate the binary perception-success judgment procedure, we
compare GPT-5.2 judge labels against human annotations for all
1{,}045 descriptions produced by the five models evaluated on
DiagVRIQ (5 models $\times$ 209 items): LLaVA-1.6-Mistral-7B,
Qwen2.5-VL-7B, GPT-4o, Claude Opus 5, and GPT-5.6 Sol. Human
annotations use the same criterion as the judge prompt: whether the
generated description correctly identifies the visual entities,
panel structure, and answer options present in the image.

\begin{table}[h]
\centering
\small
\caption{Human--judge agreement on the perception-success criterion.}
\label{tab:calibration}
\begin{tabular}{lcc}
\toprule
Model & $N$ & Agreement (\%) \\
\midrule
LLaVA-1.6-Mistral-7B & 209 & 97.1 \\
Qwen2.5-VL-7B        & 209 & 89.0 \\
GPT-4o               & 209 & 92.3 \\
Claude Opus 5        & 209 & 93.8 \\
GPT-5.6 Sol          & 209 & 91.9 \\
\midrule
Combined             & 1,045 & 92.8 \\
\bottomrule
\end{tabular}
\end{table}

Agreement is consistently high across the five model output sets,
ranging from 89.0\% to 97.1\%. Across all 1,045 descriptions, pooled
human--judge agreement is 92.8\%, with Cohen's $\kappa = 0.86$,
indicating strong agreement beyond chance.

\paragraph{Cross-family judge.}
Because GPT-5.2 also drafted the initial reference descriptions, a
judge from the same model family could share systematic biases with
the reference-generation process. We therefore use Claude Opus 4.8
as an independent cross-family judge on the 1{,}045 descriptions
from LLaVA-1.6-Mistral-7B, Qwen2.5-VL-7B, GPT-4o, Claude Opus 5, and GPT-5.6 Sol. Claude Opus
4.8 re-labels these descriptions using the same criterion while blind
to the GPT-5.2 labels. The two judges agree on 97.4\% of labels.

\section{Qualitative Examples}
\label{app:qualitative_examples}

\begin{figure*}[t]
    \centering
    \includegraphics[width=\textwidth]{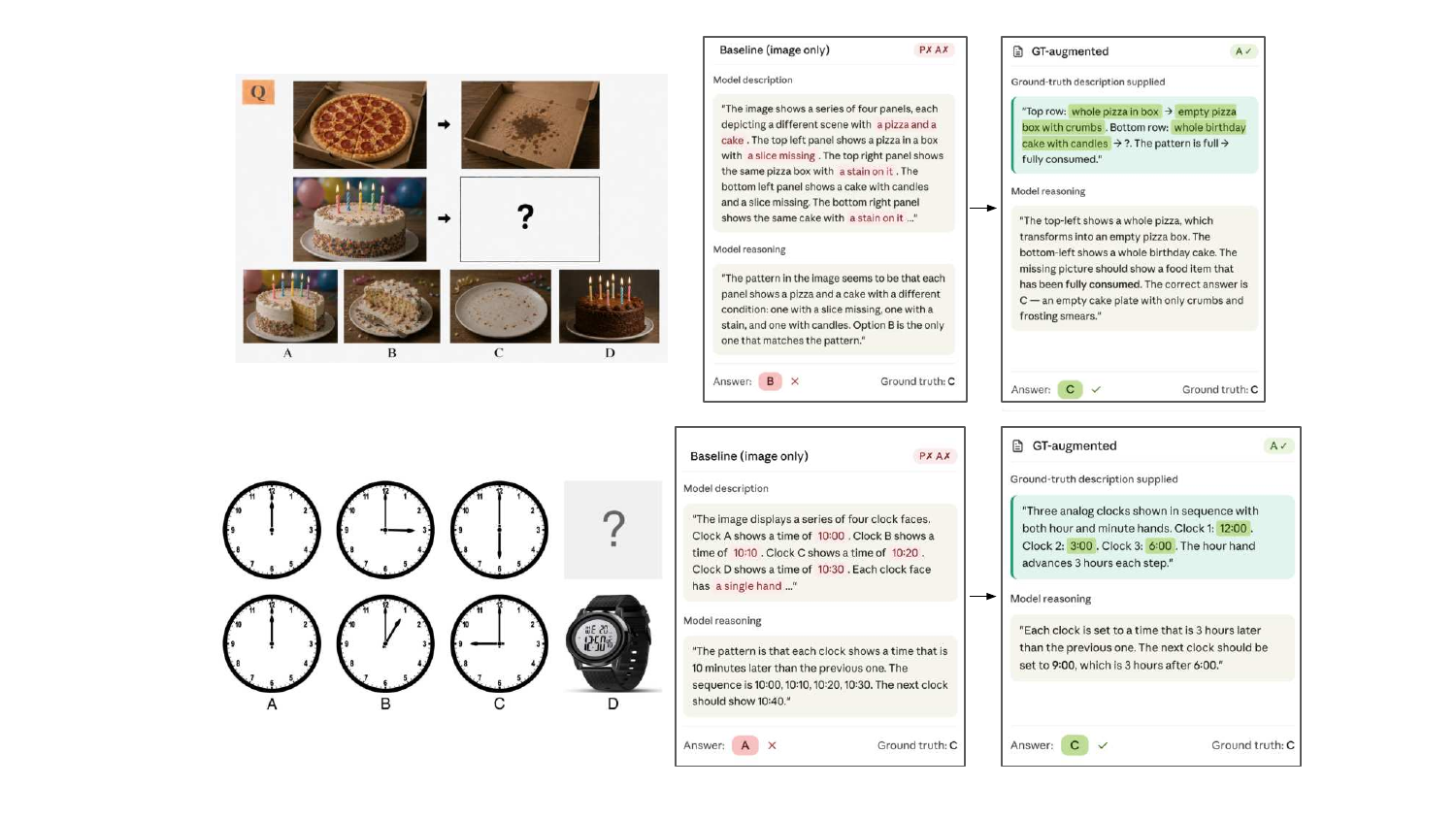}
    \caption{Representative LLaVA-1.6 examples in which the baseline response contains an incorrect visual description and an incorrect answer, while supplying the verified ground-truth description changes the final answer to the correct option.}
    \label{fig:llava-gt-fix-examples}
\end{figure*}

\begin{figure*}[t]
    \centering
    \includegraphics[width=0.7\textwidth]{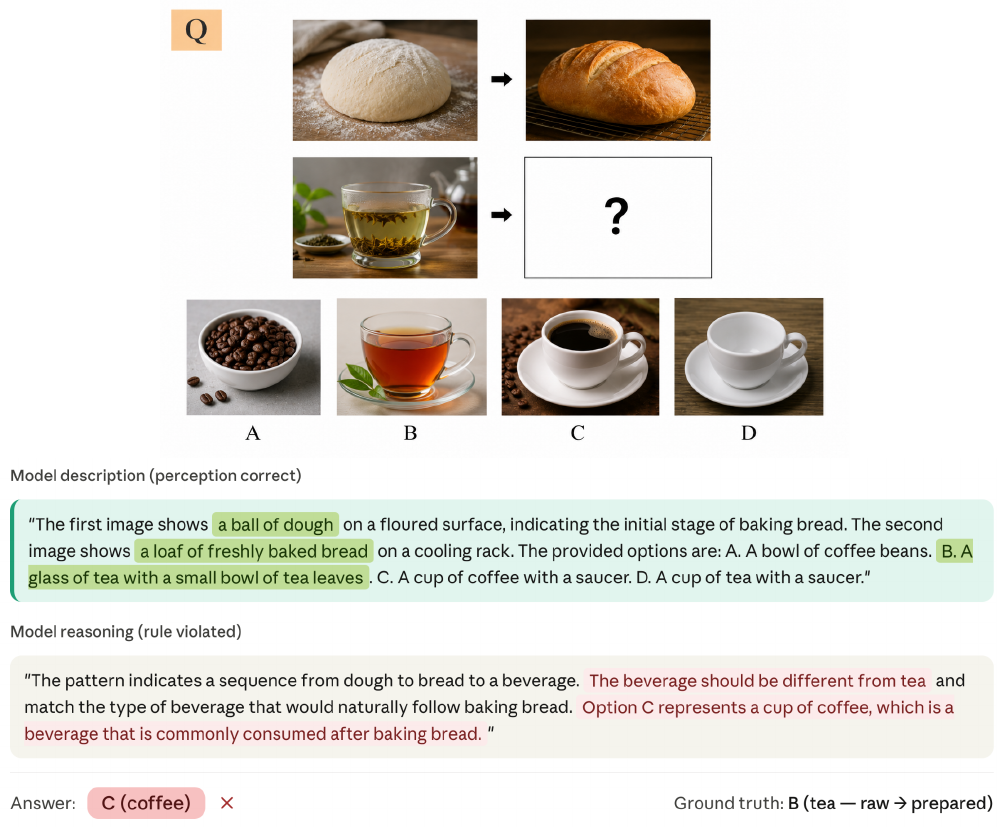}
    \caption{Representative residual reasoning failure from Qwen2.5-VL-7B. The model correctly identifies the visual entities and panel structure but applies an incorrect relational rule, producing the wrong answer even after GT augmentation.}
    \label{fig:qwen-residual-1}
\end{figure*}

\begin{figure*}[t]
    \centering
    \includegraphics[width=0.9\textwidth]{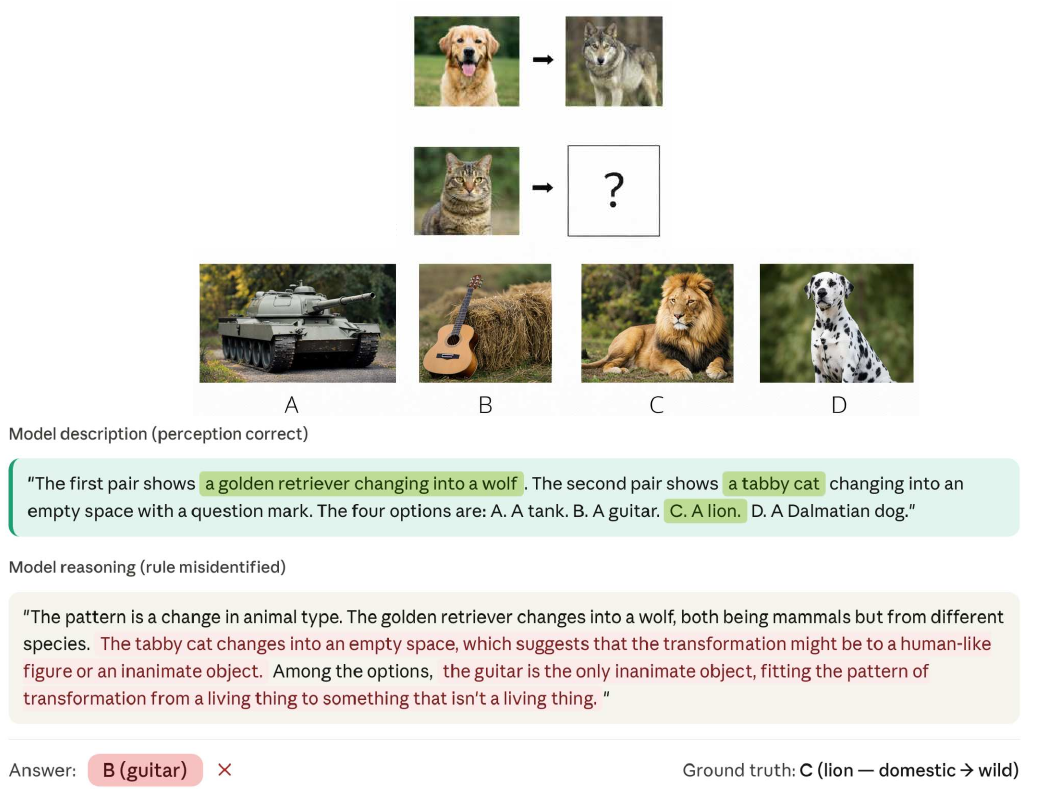}
    \caption{Additional residual reasoning failure under GT augmentation. The model correctly identifies all entities but constructs an invalid grouping strategy that cannot uniquely determine the odd-one-out item.}
    \label{fig:qwen-residual-2}
\end{figure*}

\autoref{fig:llava-gt-fix-examples}, \autoref{fig:qwen-residual-1}, and \autoref{fig:qwen-residual-2} present representative qualitative examples from the description-based attribution and GT-augmentation experiments.

\section{Per-Category DiagVRIQ Results}
\label{app:diag-per-category}

\autoref{tab:perc_reas_acc_updated} reports DiagVRIQ end-to-end
accuracy, description success, and answer accuracy conditional on a
correct description for each category.

\begin{table}[t]
\centering
\small
\caption{Per-category DiagVRIQ results (counts and \%).
Description success is the fraction of generated descriptions that
satisfy the perception-success criterion. Acc.\ $\mid P^{\checkmark}$
denotes answer accuracy conditional on the generated description
satisfying that criterion. All quantities are computed from the same
describe-then-solve generation.}
\label{tab:perc_reas_acc_updated}
\setlength{\tabcolsep}{5pt}

\begin{tabular}{llccc}
\toprule
Model & Cat. & End-to-end & \makecell{Description\\success}
& Acc.\ $\mid P^{\checkmark}$ \\
\midrule

\multirow{3}{*}{LLaVA-v1.6-Mistral-7B}
& MP (98) & 21/98 (21.4) & 6/98 (6.1) & 2/6 (33.3) \\
& OO (51) & 20/51 (39.2) & 17/51 (33.3) & 10/17 (58.8) \\
& SC (60) & 15/60 (25.0) & 2/60 (3.3) & 1/2 (50.0) \\
\midrule

\multirow{3}{*}{Qwen2.5-VL-7B}
& MP (98) & 58/98 (59.2) & 69/98 (70.4) & 51/69 (73.9) \\
& OO (51) & 27/51 (52.9) & 43/51 (84.3) & 26/43 (60.5) \\
& SC (60) & 23/60 (38.3) & 18/60 (30.0) & 13/18 (72.2) \\
\midrule

\multirow{3}{*}{GPT-4o}
& MP (98) & 86/98 (87.8) & 93/98 (94.9) & 85/93 (91.4) \\
& OO (51) & 43/51 (84.3) & 48/51 (94.1) & 43/48 (89.6) \\
& SC (60) & 48/60 (80.0) & 52/60 (86.7) & 47/52 (90.4) \\
\midrule

\multirow{3}{*}{Claude Opus 5}
& MP (98) & 89/98 (90.8) & 86/98 (87.8) & 77/86 (89.5) \\
& OO (51) & 47/51 (92.2) & 48/51 (94.1) & 45/48 (93.8) \\
& SC (60) & 54/60 (90.0) & 45/60 (75.0) & 41/45 (91.1) \\
\midrule

\multirow{3}{*}{GPT-5.6 Sol}
& MP (98) & 86/98 (87.8) & 79/98 (80.6) & 70/79 (88.6) \\
& OO (51) & 47/51 (92.2) & 48/51 (94.1) & 44/48 (91.7) \\
& SC (60) & 51/60 (85.0) & 50/60 (83.3) & 42/50 (84.0) \\
\bottomrule

\end{tabular}
\end{table}

\section{Confidence Intervals for End-to-End Accuracy}
\label{app:e2e-ci}

\autoref{tab:abstract_ci} and \autoref{tab:natural_ci} report 95\%
Wilson confidence intervals for every model--category cell in
\autoref{tab:abstract_main} and \autoref{tab:natural_main}.

\begin{table*}[t]
\centering
\footnotesize
\caption{Abstract-domain accuracy (\%) on VRIQ with 95\% Wilson confidence intervals. Category sizes are 210, 198, 141, 106, 133 for 3D, FR, MP, OO, and SC, respectively. Each item is evaluated once. Point estimates match \autoref{tab:abstract_main}.}
\label{tab:abstract_ci}
\setlength{\tabcolsep}{4pt}
\begin{tabular}{lccccc}
\toprule
Model & 3D & FR & MP & OO & SC \\
\midrule
\multicolumn{6}{l}{\textit{Open-weight}} \\
Qwen2.5-VL-3B-Instruct & 28.1 {\scriptsize[22.5, 34.5]} & 23.2 {\scriptsize[17.9, 29.6]} & 17.0 {\scriptsize[11.7, 24.1]} & 25.5 {\scriptsize[18.1, 34.5]} & 20.3 {\scriptsize[14.3, 27.9]} \\
Qwen2.5-VL-3B-Instruct-AWQ & 30.0 {\scriptsize[24.2, 36.5]} & 17.2 {\scriptsize[12.6, 23.0]} & 19.1 {\scriptsize[13.5, 26.4]} & 19.8 {\scriptsize[13.3, 28.4]} & 20.3 {\scriptsize[14.3, 27.9]} \\
Qwen2.5-VL-7B-Instruct & 32.9 {\scriptsize[26.9, 39.5]} & 25.3 {\scriptsize[19.7, 31.7]} & 31.2 {\scriptsize[24.1, 39.3]} & 26.4 {\scriptsize[19.0, 35.5]} & 40.6 {\scriptsize[32.6, 49.1]} \\
InternVL3-9B & 31.9 {\scriptsize[26.0, 38.5]} & 20.2 {\scriptsize[15.2, 26.3]} & 31.2 {\scriptsize[24.1, 39.3]} & 16.0 {\scriptsize[10.3, 24.2]} & 24.1 {\scriptsize[17.6, 32.0]} \\
LLaVA-v1.6-Mistral-7B & 28.6 {\scriptsize[22.9, 35.0]} & 20.2 {\scriptsize[15.2, 26.3]} & 19.9 {\scriptsize[14.1, 27.2]} & 27.4 {\scriptsize[19.8, 36.5]} & 15.8 {\scriptsize[10.6, 22.9]} \\
LLaVA-v1.6-Vicuna-7B & 23.8 {\scriptsize[18.6, 30.0]} & 16.2 {\scriptsize[11.7, 21.9]} & 18.4 {\scriptsize[12.9, 25.6]} & 15.1 {\scriptsize[9.5, 23.1]} & 23.3 {\scriptsize[16.9, 31.2]} \\
LLaVA-OneVision-1.5-8B-RL & 31.9 {\scriptsize[26.0, 38.5]} & 22.2 {\scriptsize[17.0, 28.5]} & 41.8 {\scriptsize[34.0, 50.1]} & 24.5 {\scriptsize[17.3, 33.5]} & 14.3 {\scriptsize[9.3, 21.2]} \\
Bee-8B-RL & 21.9 {\scriptsize[16.8, 28.0]} & 29.8 {\scriptsize[23.9, 36.5]} & 41.8 {\scriptsize[34.0, 50.1]} & 24.5 {\scriptsize[17.3, 33.5]} & 30.8 {\scriptsize[23.6, 39.1]} \\
Qwen3-VL-32B-Thinking & 33.8 {\scriptsize[27.8, 40.4]} & 31.3 {\scriptsize[25.3, 38.1]} & 34.0 {\scriptsize[26.7, 42.2]} & 34.0 {\scriptsize[25.6, 43.4]} & 44.4 {\scriptsize[36.2, 52.8]} \\
LLaVA-v1.6-34B & 22.9 {\scriptsize[17.7, 29.0]} & 20.2 {\scriptsize[15.2, 26.3]} & 26.2 {\scriptsize[19.7, 34.1]} & 19.8 {\scriptsize[13.3, 28.4]} & 24.8 {\scriptsize[18.2, 32.8]} \\
\midrule
\multicolumn{6}{l}{\textit{Proprietary}} \\
GPT-4o & 31.0 {\scriptsize[25.1, 37.5]} & 31.8 {\scriptsize[25.7, 38.6]} & 29.8 {\scriptsize[22.9, 37.8]} & 30.2 {\scriptsize[22.3, 39.5]} & 28.6 {\scriptsize[21.6, 36.8]} \\
GPT-4o-mini & 24.8 {\scriptsize[19.4, 31.0]} & 25.3 {\scriptsize[19.7, 31.7]} & 23.4 {\scriptsize[17.2, 31.0]} & 28.3 {\scriptsize[20.6, 37.5]} & 23.3 {\scriptsize[16.9, 31.2]} \\
GPT-5-mini & 21.9 {\scriptsize[16.8, 28.0]} & 24.2 {\scriptsize[18.8, 30.7]} & 25.5 {\scriptsize[19.1, 33.3]} & 26.4 {\scriptsize[19.0, 35.5]} & 24.8 {\scriptsize[18.2, 32.8]} \\
GPT-5.1 & 27.1 {\scriptsize[21.6, 33.5]} & 30.8 {\scriptsize[24.8, 37.6]} & 27.0 {\scriptsize[20.3, 34.8]} & 34.0 {\scriptsize[25.6, 43.4]} & 27.8 {\scriptsize[20.9, 36.0]} \\
Gemini-2.5-Pro & 29.5 {\scriptsize[23.8, 36.0]} & 31.3 {\scriptsize[25.3, 38.1]} & 27.7 {\scriptsize[20.9, 35.6]} & 27.4 {\scriptsize[19.8, 36.5]} & 27.1 {\scriptsize[20.2, 35.2]} \\
Claude Sonnet 4.6 & 21.9 {\scriptsize[16.8, 28.0]} & 29.8 {\scriptsize[23.9, 36.5]} & 32.6 {\scriptsize[25.4, 40.7]} & 22.6 {\scriptsize[15.7, 31.5]} & 24.1 {\scriptsize[17.6, 32.0]} \\
GPT-5.2 & 36.2 {\scriptsize[30.0, 42.9]} & 39.9 {\scriptsize[33.3, 46.8]} & 44.0 {\scriptsize[36.0, 52.2]} & 32.1 {\scriptsize[24.0, 41.5]} & 25.6 {\scriptsize[18.9, 33.6]} \\
\midrule
\multicolumn{6}{l}{\textit{2026 frontier}} \\
Gemini 3.1 Pro Preview & 35.2 {\scriptsize[29.1, 41.9]} & 55.1 {\scriptsize[48.1, 61.8]} & 60.3 {\scriptsize[52.0, 68.0]} & 30.2 {\scriptsize[22.3, 39.5]} & 25.6 {\scriptsize[18.9, 33.6]} \\
Claude Opus 5 & 23.8 {\scriptsize[18.6, 30.0]} & 76.8 {\scriptsize[70.4, 82.1]} & 46.1 {\scriptsize[38.1, 54.3]} & 60.4 {\scriptsize[50.9, 69.2]} & 27.8 {\scriptsize[20.9, 36.0]} \\
GPT-5.6 Sol (medium) & 50.0 {\scriptsize[43.3, 56.7]} & 84.8 {\scriptsize[79.2, 89.2]} & 85.1 {\scriptsize[78.3, 90.0]} & 51.9 {\scriptsize[42.5, 61.2]} & 64.7 {\scriptsize[56.2, 72.3]} \\
GPT-5.6 Sol (high) & 44.8 {\scriptsize[38.2, 51.5]} & 84.8 {\scriptsize[79.2, 89.2]} & 80.1 {\scriptsize[72.8, 85.9]} & 65.1 {\scriptsize[55.6, 73.5]} & 79.7 {\scriptsize[72.1, 85.7]} \\
\bottomrule
\end{tabular}
\end{table*}

\begin{table*}[t]
\centering
\footnotesize
\caption{Natural-domain accuracy (\%) on VRIQ with 95\% Wilson confidence intervals. Category sizes are 111, 97, 146, 149, 100 for 3D, FR, MP, OO, and SC, respectively. Each item is evaluated once. Point estimates match \autoref{tab:natural_main}.}
\label{tab:natural_ci}
\setlength{\tabcolsep}{4pt}
\begin{tabular}{lccccc}
\toprule
Model & 3D & FR & MP & OO & SC \\
\midrule
\multicolumn{6}{l}{\textit{Open-weight}} \\
Qwen2.5-VL-3B-Instruct & 29.7 {\scriptsize[22.0, 38.8]} & 40.2 {\scriptsize[31.0, 50.2]} & 43.2 {\scriptsize[35.4, 51.3]} & 40.9 {\scriptsize[33.4, 49.0]} & 34.0 {\scriptsize[25.5, 43.7]} \\
Qwen2.5-VL-3B-Instruct-AWQ & 25.2 {\scriptsize[18.1, 34.0]} & 35.1 {\scriptsize[26.3, 45.0]} & 28.8 {\scriptsize[22.0, 36.6]} & 36.2 {\scriptsize[29.0, 44.2]} & 37.0 {\scriptsize[28.2, 46.8]} \\
Qwen2.5-VL-7B-Instruct & 29.7 {\scriptsize[22.0, 38.8]} & 35.1 {\scriptsize[26.3, 45.0]} & 39.7 {\scriptsize[32.2, 47.8]} & 49.7 {\scriptsize[41.7, 57.6]} & 25.0 {\scriptsize[17.5, 34.3]} \\
InternVL3-9B & 39.6 {\scriptsize[31.0, 48.9]} & 45.4 {\scriptsize[35.8, 55.3]} & 34.9 {\scriptsize[27.7, 43.0]} & 40.3 {\scriptsize[32.7, 48.3]} & 40.0 {\scriptsize[30.9, 49.8]} \\
LLaVA-v1.6-Mistral-7B & 27.0 {\scriptsize[19.6, 36.0]} & 17.5 {\scriptsize[11.2, 26.3]} & 24.7 {\scriptsize[18.4, 32.2]} & 30.2 {\scriptsize[23.4, 38.0]} & 20.0 {\scriptsize[13.3, 28.9]} \\
LLaVA-v1.6-Vicuna-7B & 18.9 {\scriptsize[12.7, 27.2]} & 15.5 {\scriptsize[9.6, 24.0]} & 15.1 {\scriptsize[10.2, 21.8]} & 34.2 {\scriptsize[27.1, 42.2]} & 24.0 {\scriptsize[16.7, 33.2]} \\
LLaVA-OneVision-1.5-8B-RL & 36.9 {\scriptsize[28.5, 46.2]} & 45.4 {\scriptsize[35.8, 55.3]} & 52.1 {\scriptsize[44.0, 60.0]} & 28.9 {\scriptsize[22.2, 36.6]} & 41.0 {\scriptsize[31.9, 50.8]} \\
Bee-8B-RL & 50.5 {\scriptsize[41.3, 59.6]} & 29.9 {\scriptsize[21.7, 39.6]} & 45.9 {\scriptsize[38.0, 54.0]} & 24.8 {\scriptsize[18.6, 32.3]} & 33.0 {\scriptsize[24.6, 42.7]} \\
Qwen3-VL-32B-Thinking & 41.4 {\scriptsize[32.7, 50.7]} & 49.5 {\scriptsize[39.7, 59.3]} & 50.0 {\scriptsize[42.0, 58.0]} & 53.7 {\scriptsize[45.7, 61.5]} & 42.0 {\scriptsize[32.8, 51.8]} \\
LLaVA-v1.6-34B & 27.0 {\scriptsize[19.6, 36.0]} & 19.6 {\scriptsize[12.9, 28.6]} & 43.2 {\scriptsize[35.4, 51.3]} & 44.3 {\scriptsize[36.6, 52.3]} & 40.0 {\scriptsize[30.9, 49.8]} \\
\midrule
\multicolumn{6}{l}{\textit{Proprietary}} \\
GPT-4o & 47.7 {\scriptsize[38.7, 57.0]} & 40.2 {\scriptsize[31.0, 50.2]} & 62.3 {\scriptsize[54.2, 69.8]} & 57.0 {\scriptsize[49.0, 64.7]} & 64.0 {\scriptsize[54.2, 72.7]} \\
GPT-4o-mini & 38.7 {\scriptsize[30.2, 48.0]} & 40.2 {\scriptsize[31.0, 50.2]} & 47.9 {\scriptsize[40.0, 56.0]} & 48.3 {\scriptsize[40.4, 56.3]} & 44.0 {\scriptsize[34.7, 53.8]} \\
GPT-5-mini & 27.9 {\scriptsize[20.4, 36.9]} & 40.2 {\scriptsize[31.0, 50.2]} & 50.7 {\scriptsize[42.7, 58.7]} & 48.3 {\scriptsize[40.4, 56.3]} & 49.0 {\scriptsize[39.4, 58.7]} \\
GPT-5.1 & 41.4 {\scriptsize[32.7, 50.7]} & 40.2 {\scriptsize[31.0, 50.2]} & 58.9 {\scriptsize[50.8, 66.6]} & 59.7 {\scriptsize[51.7, 67.3]} & 59.0 {\scriptsize[49.2, 68.1]} \\
Gemini-2.5-Pro & 58.6 {\scriptsize[49.3, 67.3]} & 33.0 {\scriptsize[24.4, 42.8]} & 65.1 {\scriptsize[57.0, 72.3]} & 49.7 {\scriptsize[41.7, 57.6]} & 61.0 {\scriptsize[51.2, 70.0]} \\
Claude Sonnet 4.6 & 44.1 {\scriptsize[35.3, 53.4]} & 25.8 {\scriptsize[18.1, 35.3]} & 63.7 {\scriptsize[55.6, 71.1]} & 55.7 {\scriptsize[47.7, 63.4]} & 72.0 {\scriptsize[62.5, 79.9]} \\
GPT-5.2 & 39.6 {\scriptsize[31.0, 48.9]} & 46.4 {\scriptsize[36.8, 56.3]} & 56.2 {\scriptsize[48.1, 64.0]} & 51.7 {\scriptsize[43.7, 59.6]} & 68.0 {\scriptsize[58.3, 76.3]} \\
\midrule
\multicolumn{6}{l}{\textit{2026 frontier}} \\
Gemini 3.1 Pro Preview & 55.0 {\scriptsize[45.7, 63.9]} & 80.4 {\scriptsize[71.4, 87.1]} & 89.7 {\scriptsize[83.7, 93.7]} & 96.0 {\scriptsize[91.5, 98.1]} & 70.0 {\scriptsize[60.4, 78.1]} \\
Claude Opus 5 & 50.5 {\scriptsize[41.3, 59.6]} & 74.2 {\scriptsize[64.7, 81.9]} & 86.3 {\scriptsize[79.8, 91.0]} & 89.9 {\scriptsize[84.1, 93.8]} & 82.0 {\scriptsize[73.3, 88.3]} \\
GPT-5.6 Sol (medium) & 60.4 {\scriptsize[51.1, 69.0]} & 35.1 {\scriptsize[26.3, 45.0]} & 80.1 {\scriptsize[72.9, 85.8]} & 79.9 {\scriptsize[72.7, 85.5]} & 75.0 {\scriptsize[65.7, 82.5]} \\
GPT-5.6 Sol (high) & 55.0 {\scriptsize[45.7, 63.9]} & 35.1 {\scriptsize[26.3, 45.0]} & 80.1 {\scriptsize[72.9, 85.8]} & 79.9 {\scriptsize[72.7, 85.5]} & 85.0 {\scriptsize[76.7, 90.7]} \\
\bottomrule
\end{tabular}
\end{table*}

\section{Ground-Truth Augmentation Prompt}
\label{appendix:gt_prompt}

The GT-augmented condition (\autoref{sec:reasoning-with-gt}) supplies the model with the verified ground-truth description alongside the image and question. We use the prompt in \autoref{fig:gt_prompt} for all evaluated VLMs.

\begin{figure}[h]
\centering
\begin{minipage}{\columnwidth}
\small
\begin{tcolorbox}[colback=gray!5, colframe=gray!50, boxrule=0.5pt]
\textbf{System.} You are an expert at solving visual reasoning puzzles. You will be shown an image, a verified textual description of what the image contains, and a multiple-choice question. Use both the image and the description when reasoning. The description has been prepared by human annotators and is reliable.

\medskip
\textbf{User.}\\
\texttt{[image]}\\
\textbf{Description:} \texttt{\{ground\_truth\_description\}}\\
\textbf{Question:} \texttt{\{question\}}\\
\textbf{Options:} A) \texttt{\{a\}} \quad B) \texttt{\{b\}} \quad C) \texttt{\{c\}} \quad D) \texttt{\{d\}}

\medskip
Think step by step using both the image and the description. Then provide your final answer as a single letter (A, B, C, or D).
\end{tcolorbox}
\end{minipage}
\caption{Prompt used for the GT-augmented condition in
\autoref{sec:reasoning-with-gt}.}
\label{fig:gt_prompt}
\end{figure}

\section{\texttt{o3} Tool-Use Examples}
\label{app:o3-traces}

We provide three representative examples demonstrating how tool-augmented reasoning performs across different task categories and failure modes.

\subsection{Example 1: Rotation-Based Abstract Task (Success)}

\begin{figure}[ht]
\centering
\includegraphics[width=0.45\linewidth]{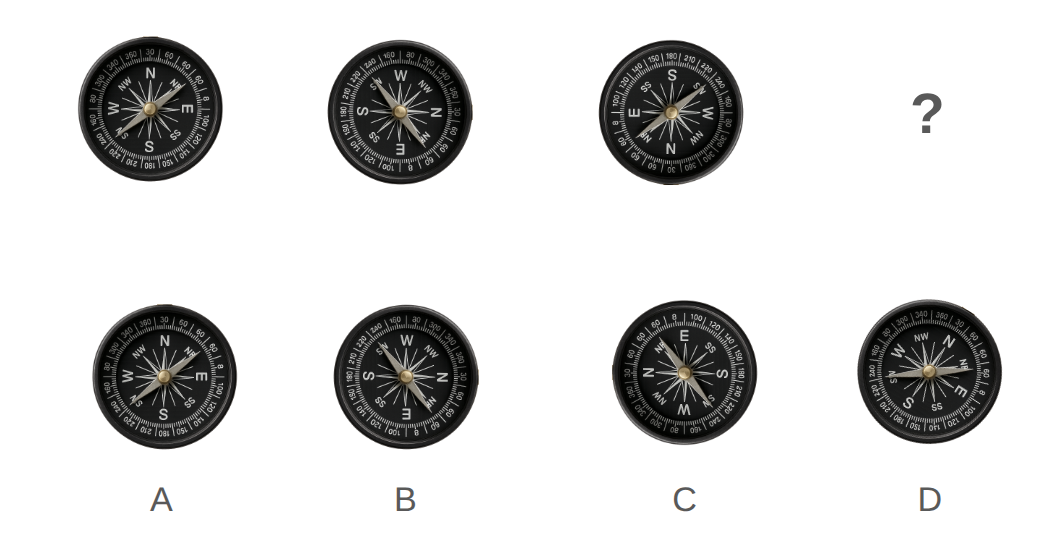}
\caption{Input sequence with the missing final image and multiple-choice options (A--D).}
\label{fig:o3_compass_input}
\end{figure}

\textbf{Task.} Identify the next image in the sequence of compass orientations shown in \autoref{fig:o3_compass_input}.

\textbf{Tool call sequence (logged):}
\begin{verbatim}
1. crop(x=40, y=20, w=240, h=240)
2. crop(x=320, y=20, w=240, h=240)
3. crop(x=600, y=20, w=240, h=240)
4. rotate(angle=90)
5. rotate(angle=180)
6. rotate(angle=270)
\end{verbatim}

\begin{figure}[ht]
\centering
\includegraphics[width=0.25\linewidth]{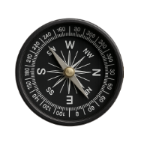}
\caption{Reference image after 270$^\circ$ rotation produced by the tool, used for comparison with the answer options.}
\label{fig:o3_compass_rotated}
\end{figure}

\begin{llmresponse}
\ttfamily
The pattern is brutally simple:

Each compass rotates 90° clockwise each step:
1st has N on top → 2nd has W on top → 3rd has S on top → next must have E on top.

Also, the missing wedge on the left grows each step, so the next one should still have that left-side cut.

Only Option C matches (E at the top + left-side cut).

Answer: C
\end{llmresponse}

\textbf{Predicted answer:} C (correct). \textbf{Ground truth:} C.

\paragraph{Why tools help.}
Instead of estimating rotation directly from the original image, \texttt{o3} explicitly generates a rotated reference (\autoref{fig:o3_compass_rotated}) and compares it against candidate options. This example illustrates how image manipulation can turn an orientation judgment into a more direct visual comparison.

\subsection{Example 2: Sequence Completion with Counting Errors (Natural Domain)}

\begin{figure}[ht]
\centering
\includegraphics[width=0.7\linewidth]{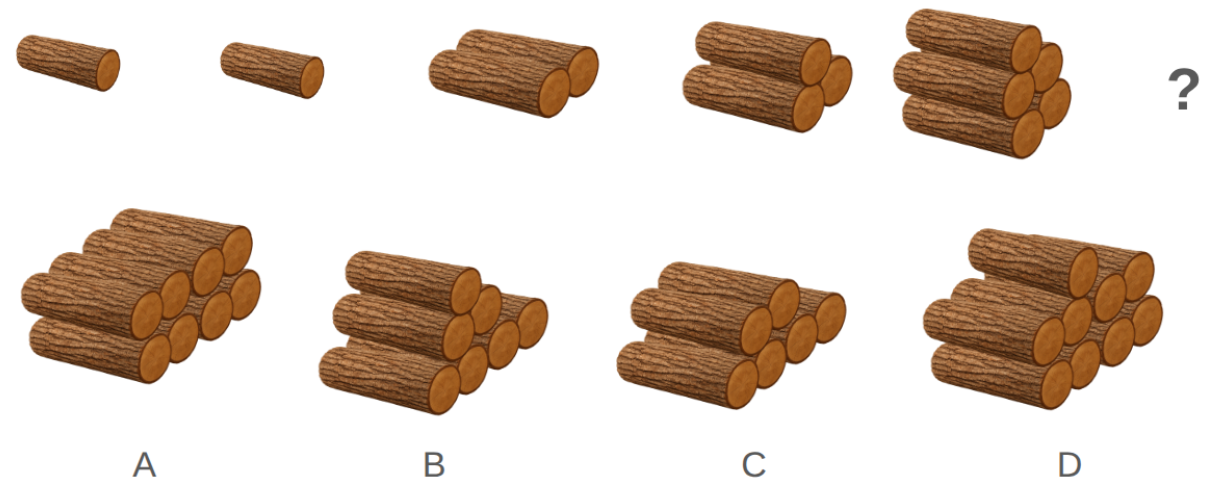}
\caption{Natural-domain sequence completion task with missing final image and answer options (A--D).}
\label{fig:o3_logs_input}
\end{figure}

\textbf{Task.} Predict the next image in the log-stacking sequence shown in \autoref{fig:o3_logs_input}.

\textbf{Tool call sequence (logged):}
\begin{verbatim}
1. crop(...)  % isolate sequence region
2. crop(...)  % option A
3. crop(...)  % option B
4. crop(...)  % option C
5. crop(...)  % option D
\end{verbatim}

\begin{figure}[ht]
\centering
\includegraphics[width=0.55\linewidth]{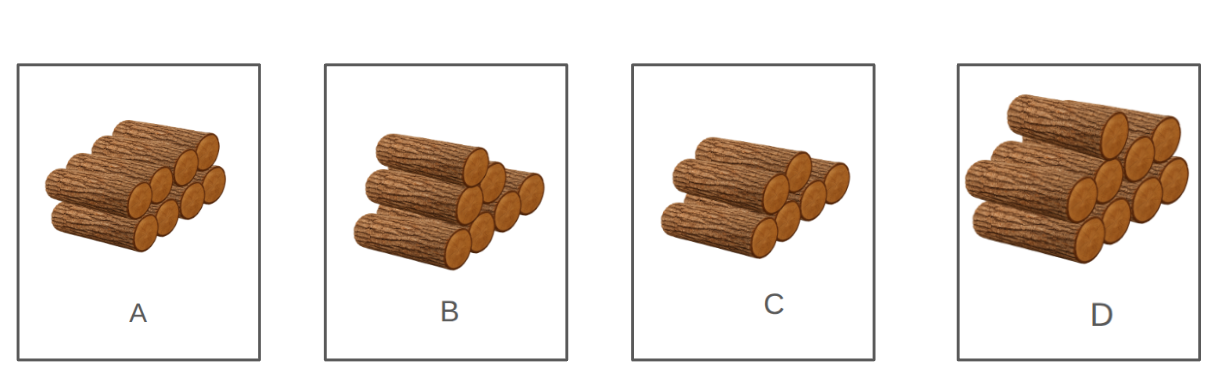}
\caption{Cropped candidate options used by \texttt{o3} for visual inspection and counting.}
\label{fig:o3_logs_crops}
\end{figure}

\begin{llmresponse}
\ttfamily
Count the logs in the top sequence:

1st: 1

2nd: 2

3rd: 2 (same total, just arranged slightly differently)

4th: 3

5th: 5

That’s the Fibonacci pattern starting as 1, 1, 2, 3, 5, so the next should be 8 logs.

Among the options, only D shows a stable stack of 8 logs (arranged as 3 + 3 + 2).

Answer: D
\end{llmresponse}

\textbf{Predicted answer:} D (incorrect). \textbf{Ground truth:} A.

\paragraph{Failure analysis.}
The tool trace reveals that \texttt{o3} inferred a plausible sequence pattern and predicted that the next configuration should contain eight logs. However, its own counts of the top sequence (1, 2, 2, 3, 5) do not match the Fibonacci pattern it then asserts (1, 1, 2, 3, 5), and after isolating the answer options (\autoref{fig:o3_logs_crops}), the model also miscounted logs in the rendered stacks because of occlusion and overlapping cylindrical structures, leading to the wrong answer. This example shows that tool-assisted isolation can improve inspection while fine-grained counting errors may still persist in cluttered natural scenes.

\subsection{Example 3: Matrix Reasoning with Persistent Errors (Abstract Domain)}

\begin{figure}[ht]
\centering
\includegraphics[width=0.5\linewidth]{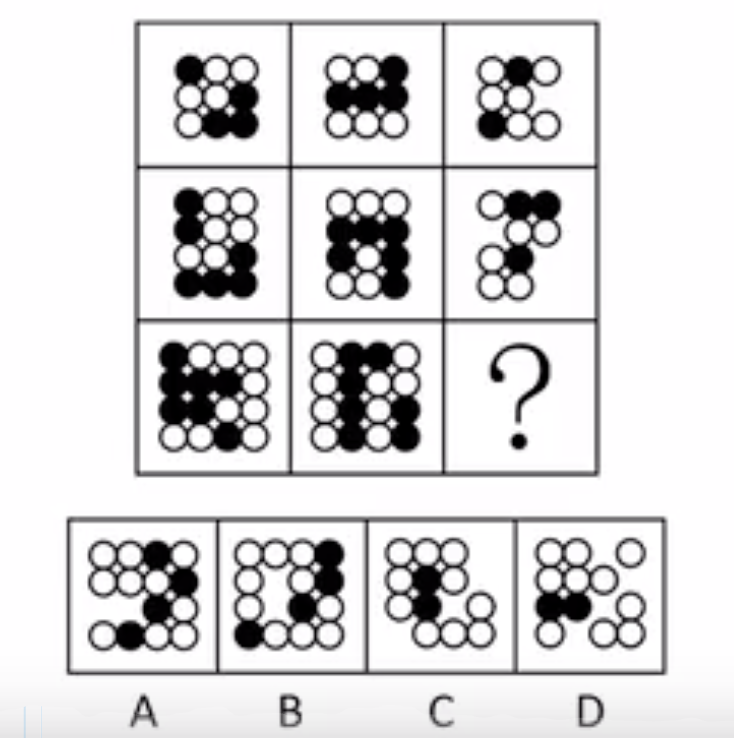}
\caption{Abstract matrix reasoning task with one missing cell (bottom-right).}
\label{fig:matrix_input}
\end{figure}

\textbf{Task.} Predict the missing cell in a 3$\times$3 matrix of circle patterns (\autoref{fig:matrix_input}).

\paragraph{Tool execution summary.}
Over 352 seconds, the model performed:
\begin{itemize}
    \item 13 cropping operations (matrix cells and answer options),
    \item 8 circle-detection attempts using HoughCircles with varying parameters,
    \item Position clustering to classify black vs. white circles.
\end{itemize}

\paragraph{What went right.}
The model correctly identified progressive grid size increases (3$\times$3 $\rightarrow$ 4$\times$3 $\rightarrow$ 5$\times$4), demonstrating successful structural pattern recognition.

\paragraph{What went wrong.}
Despite extensive tool use, perception errors persisted:
\begin{itemize}
    \item Options A and B: 14 circles detected (4 black, 10 white).
    \item Options C and D: 11 circles detected (1 black, 10 white).
    \item Cell (Row 1, Col 3): only 5 circles detected (undercount).
    \item Cell (Row 2, Col 3): 9 circles detected.
\end{itemize}
These miscounts corrupted the final decision. Additionally, the model failed to detect the actual transformation rule: circles that are black in \emph{both} columns 1 and 2 should be removed in column 3.

\begin{figure}[ht]
\centering
\includegraphics[width=0.55\linewidth]{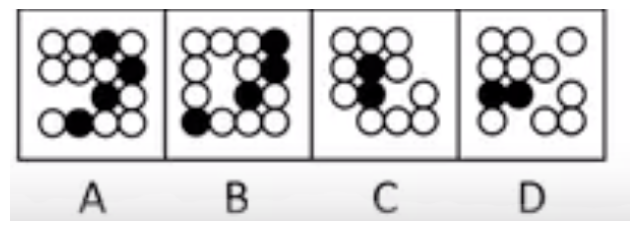}
\caption{Tool execution trace showing repeated circle-detection attempts. Despite systematic cropping, accurate enumeration and spatial localization remain unreliable under clutter.}
\label{fig:matrix_trace}
\end{figure}

\begin{llmresponse}
\ttfamily
A.
Because in the rightmost column, the black circles keep the same ``top-right pair'' pattern from row 1 → row 2, and in row 3 that pattern expands into the larger layout in the only way that stays consistent with the build-up seen in the first two columns. Option A matches that continuation; the others either drop required blacks or place them where the progression never puts them.
\end{llmresponse}

\textbf{Predicted answer:} A (incorrect). \textbf{Ground truth:} B.

\paragraph{Key takeaway.}
Tool augmentation supported systematic inspection of the matrix but did not eliminate counting and spatial-localization errors, and the model still failed to recover the correct transformation rule. The example illustrates that additional visual operations and computation can help without guaranteeing a correct solution.

\section{Dataset Samples}
\label{app:dataset-samples}

We provide additional sample items spanning the five reasoning categories and both abstract and natural domains. We also include representative DiagVRIQ items paired with their human-verified ground-truth descriptions.

\begin{figure*}[p]
    \centering
    \includegraphics[width=0.75\linewidth,height=0.85\textheight,keepaspectratio]{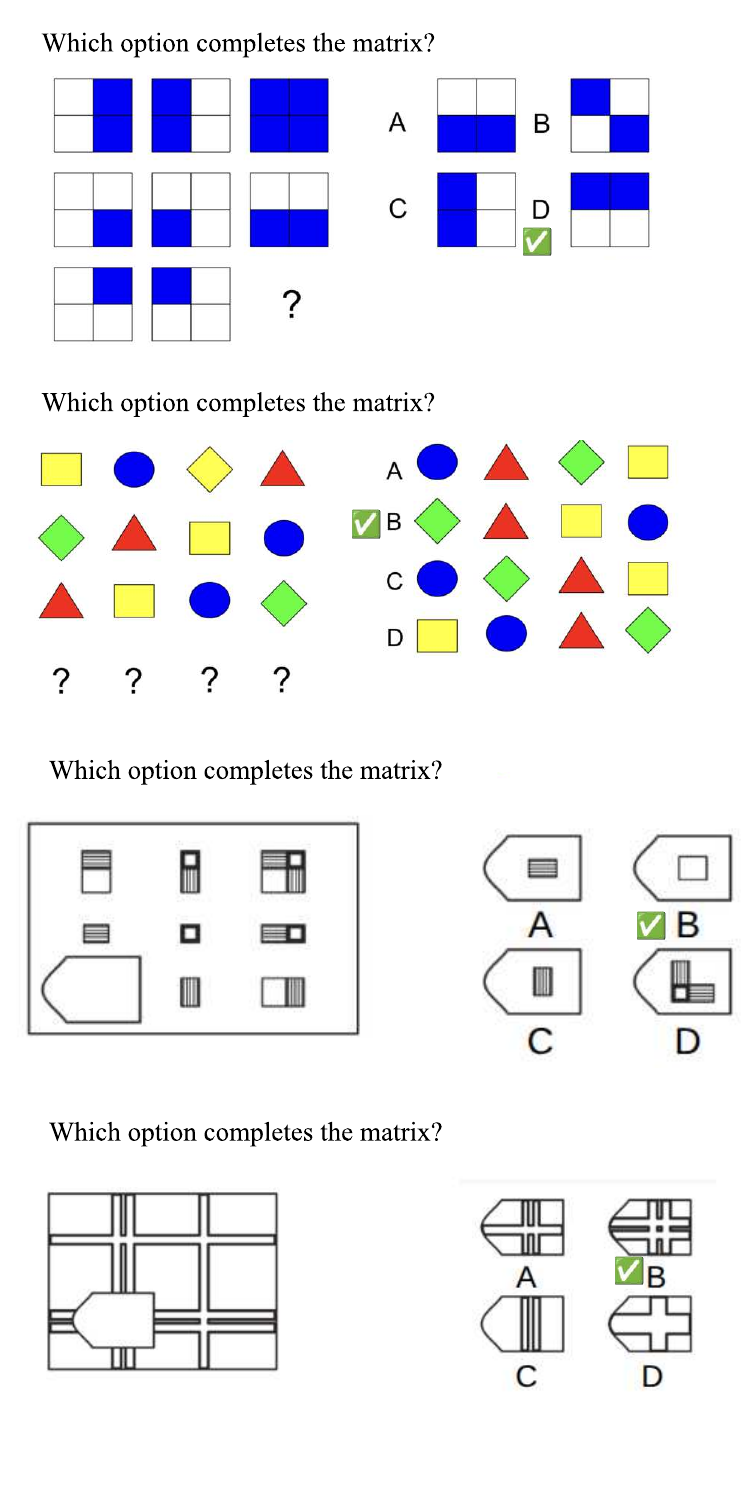}
    \caption{Sample VRIQ abstract Matrix Prediction items. Each puzzle requires identifying the missing panel in a $3 \times 3$ matrix based on a transformation rule applied across rows, columns, or both.}
    \label{fig:abstract-mp-samples}
\end{figure*}

\begin{figure*}[p]
    \centering
    \includegraphics[width=0.75\linewidth,height=0.85\textheight,keepaspectratio]{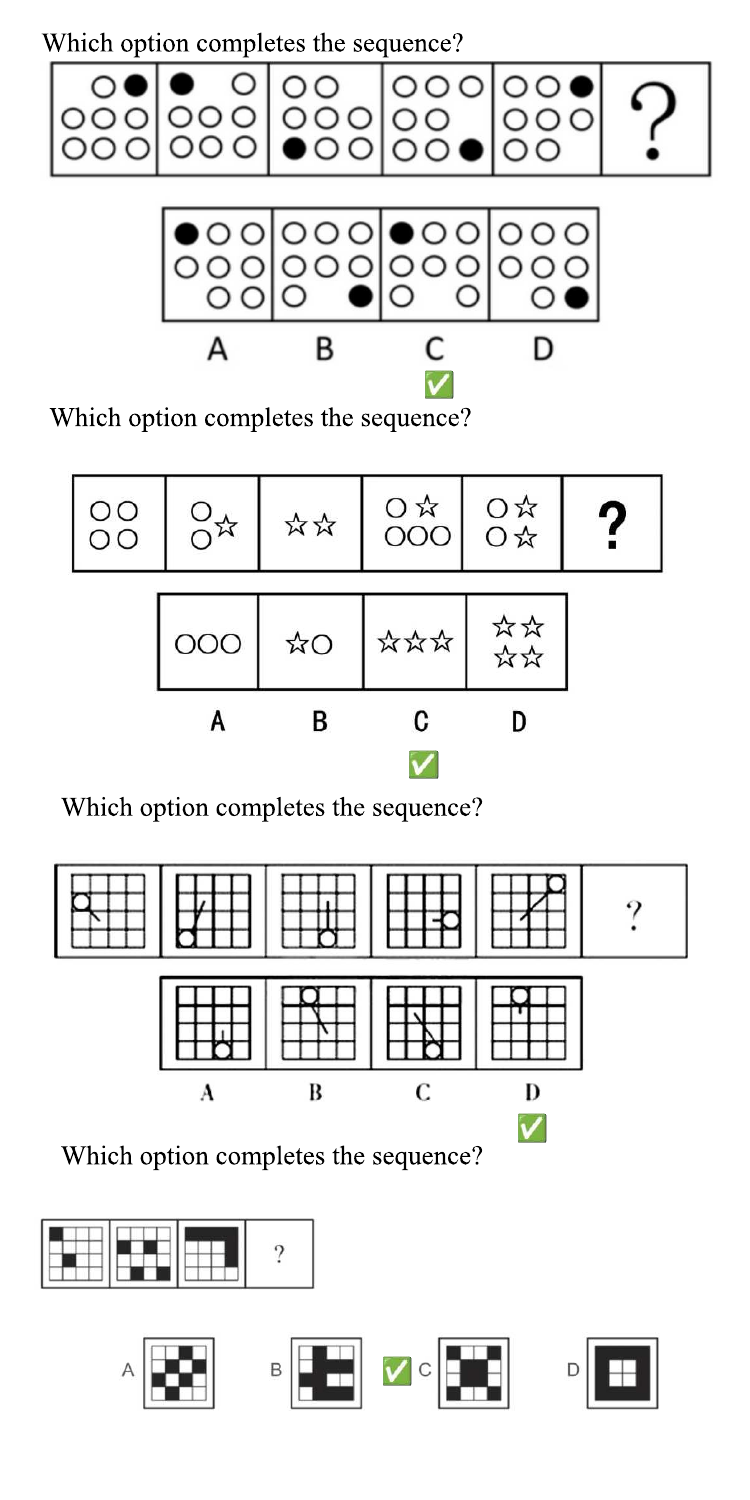}
    \caption{Sample VRIQ abstract Sequence Completion items. Each puzzle requires identifying the next element in a sequence governed by a positional, structural, or arithmetic rule.}
    \label{fig:abstract-sc-samples}
\end{figure*}

\begin{figure*}[p]
    \centering
    \includegraphics[width=0.75\linewidth,height=0.85\textheight,keepaspectratio]{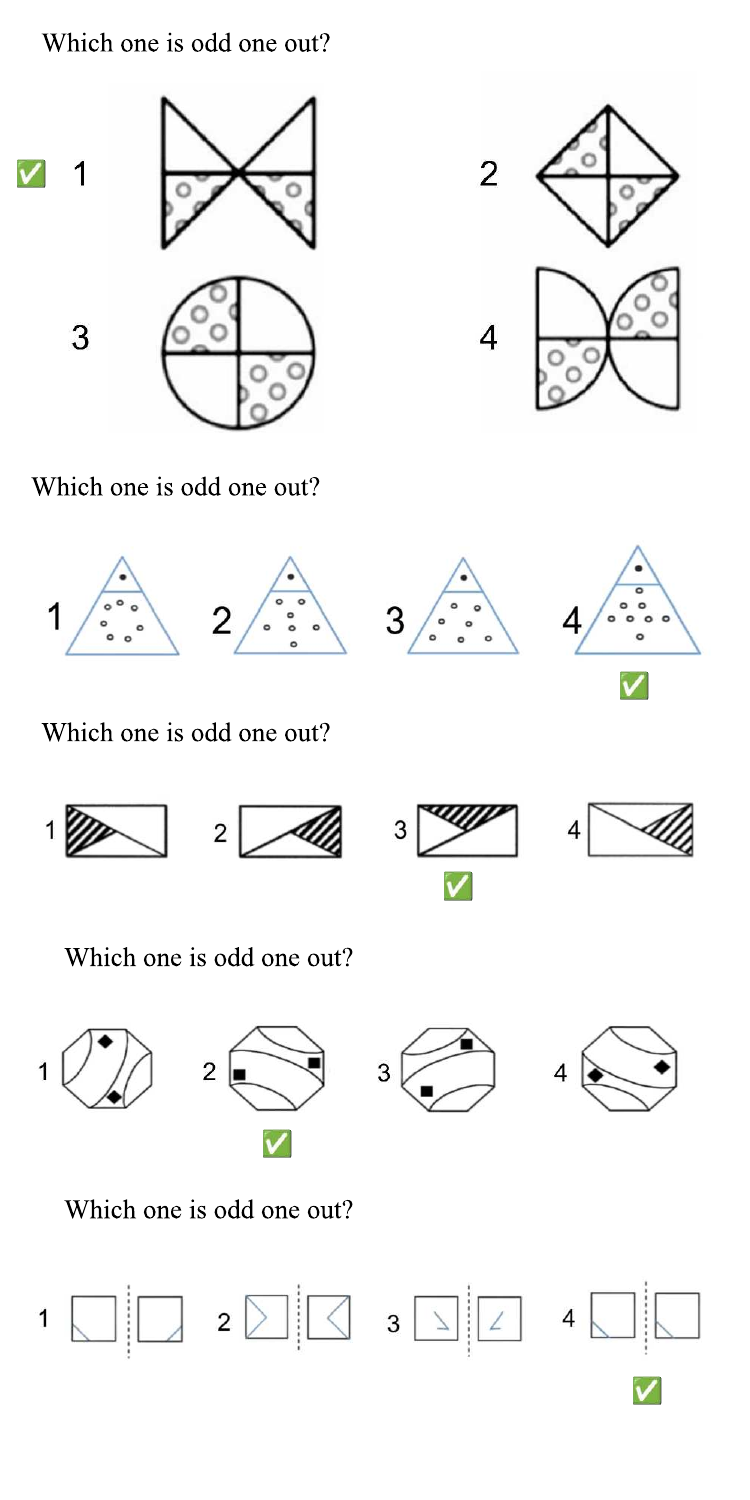}
    \caption{Sample VRIQ abstract Odd One Out items. Each puzzle requires identifying the candidate that does not share a common visual property with the others.}
    \label{fig:abstract-oo-samples}
\end{figure*}

\begin{figure*}[p]
    \centering
    \includegraphics[width=0.75\linewidth,height=0.85\textheight,keepaspectratio]{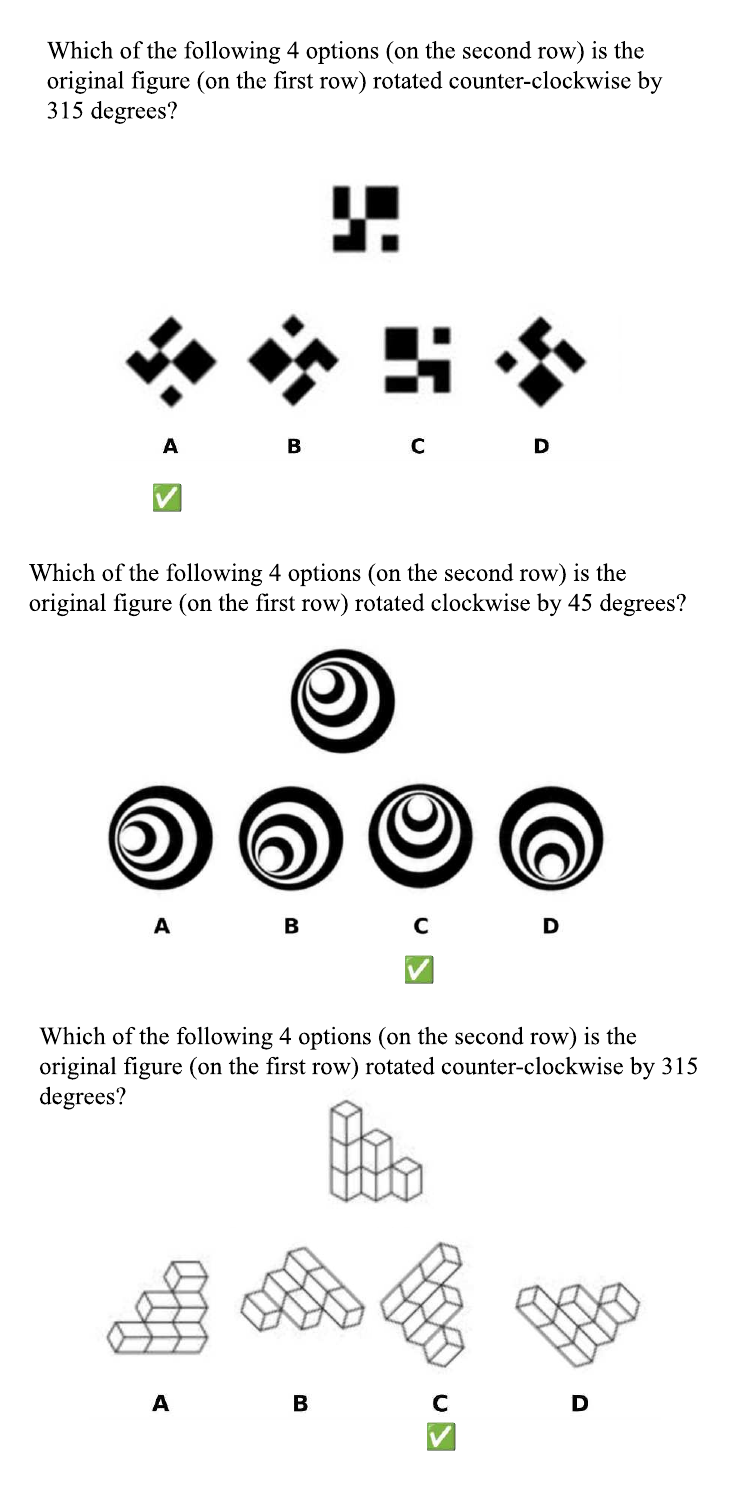}
    \caption{Sample VRIQ abstract Figure Rotation items. Each puzzle requires identifying a rotated version of a geometric figure.}
    \label{fig:abstract-fr-samples}
\end{figure*}

\begin{figure*}[p]
    \centering
    \includegraphics[width=0.95\linewidth,height=0.85\textheight,keepaspectratio]{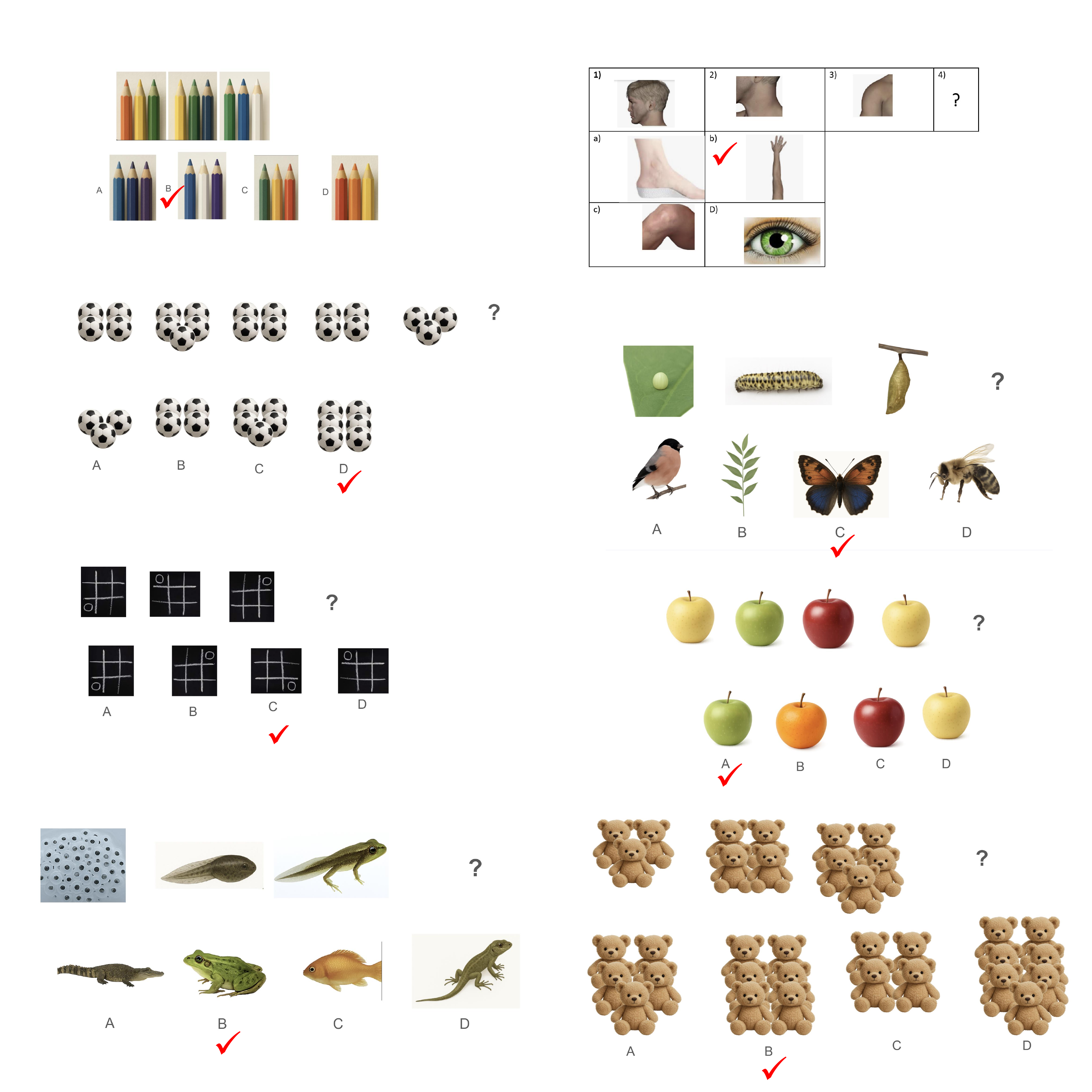}
    \caption{Sample VRIQ natural Sequence Completion questions, with the correct option marked by \checkmark.}
    \label{fig:natural-sc}
\end{figure*}

\begin{figure*}[p]
    \centering
    \includegraphics[width=0.8\linewidth,height=0.85\textheight,keepaspectratio]{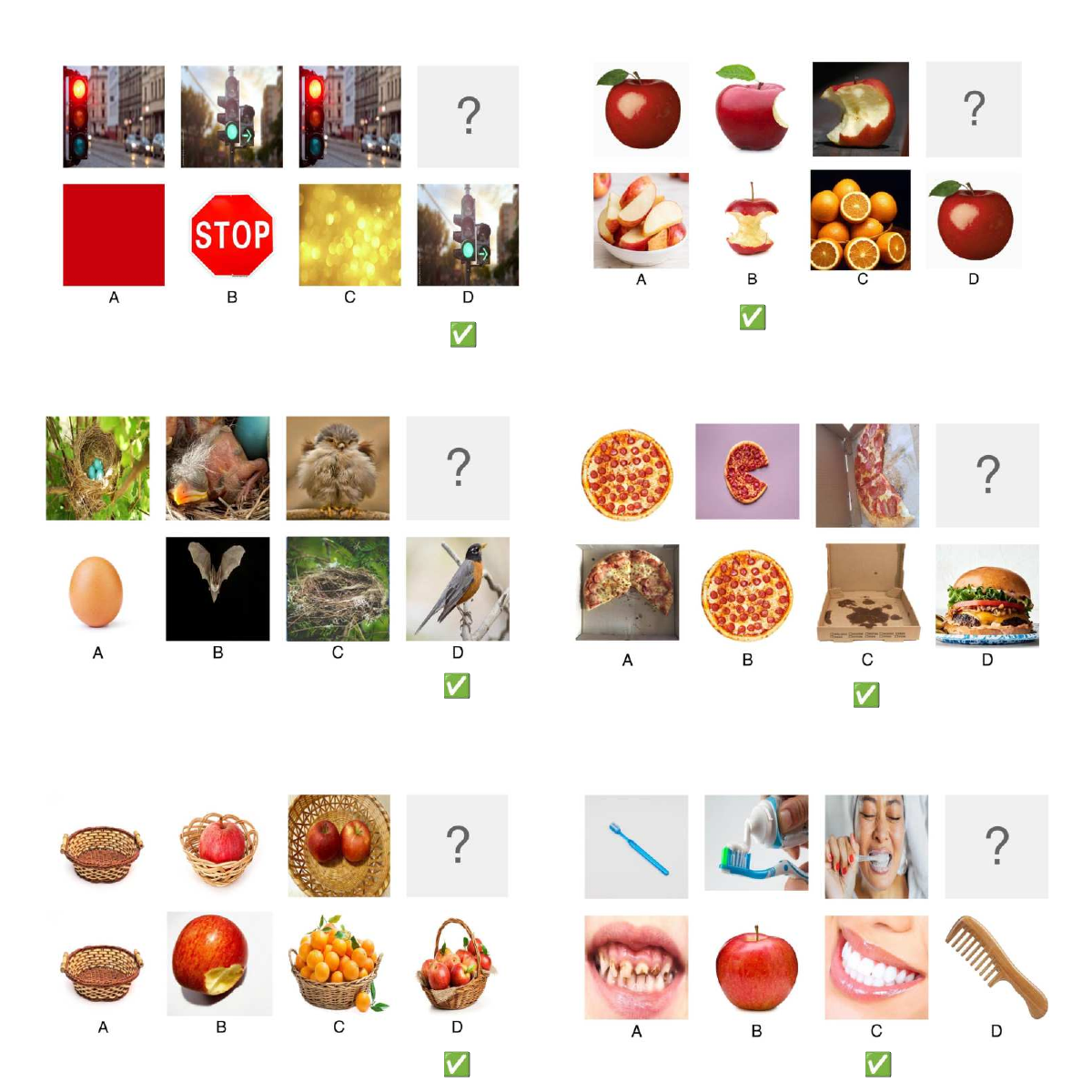}
    \caption{Sample VRIQ natural Sequence Completion items. Each puzzle requires identifying the next element in a sequence of real-world entities or scenes that follow an underlying rule.}
    \label{fig:natural-sc-samples}
\end{figure*}

\begin{figure*}[p]
    \centering
    \includegraphics[width=0.8\linewidth,height=0.85\textheight,keepaspectratio]{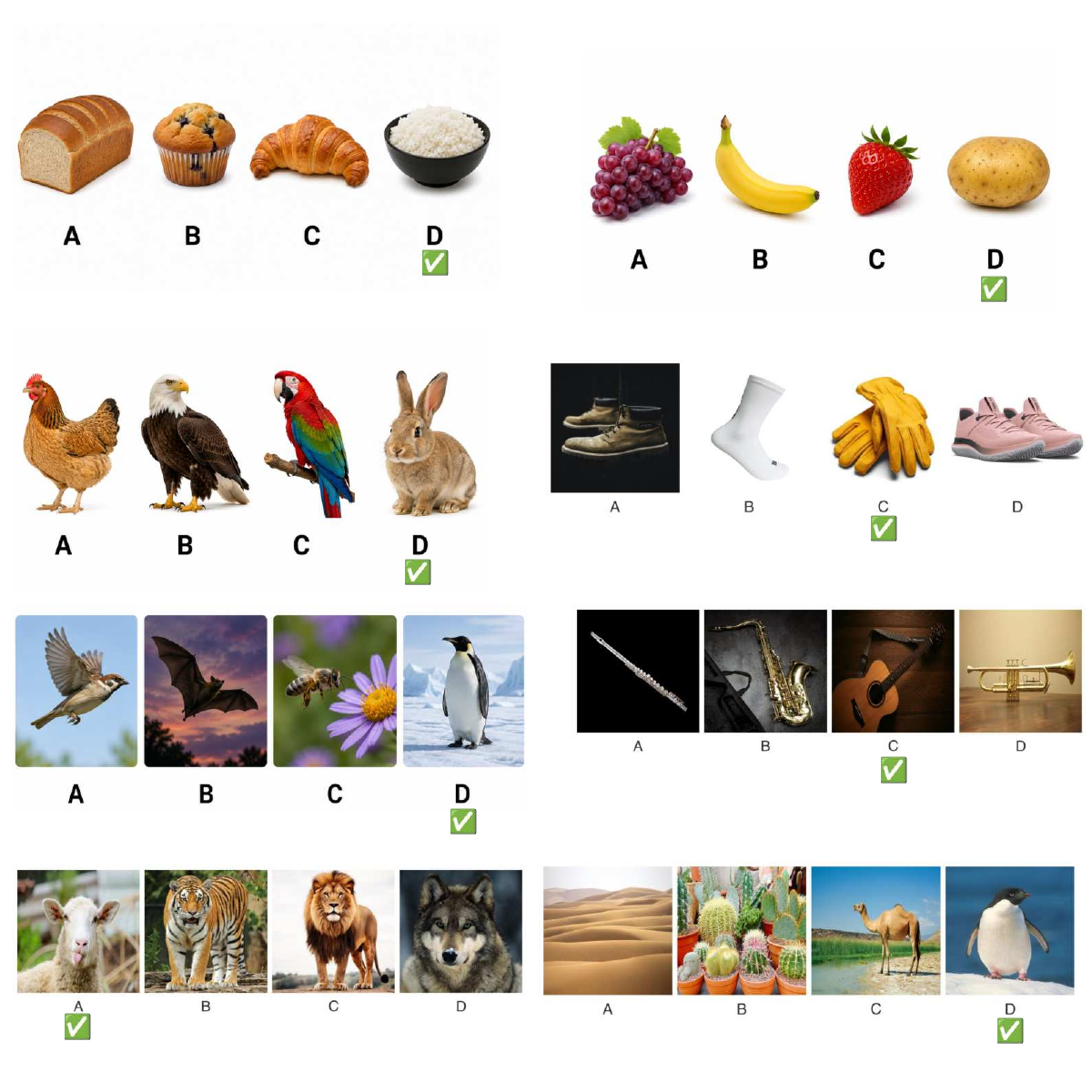}
    \caption{Sample VRIQ natural Odd One Out items. Each puzzle requires identifying the entity that does not belong to a shared category.}
    \label{fig:natural-oo-samples}
\end{figure*}

\begin{figure*}[p]
    \centering
    \includegraphics[width=0.8\linewidth,height=0.85\textheight,keepaspectratio]{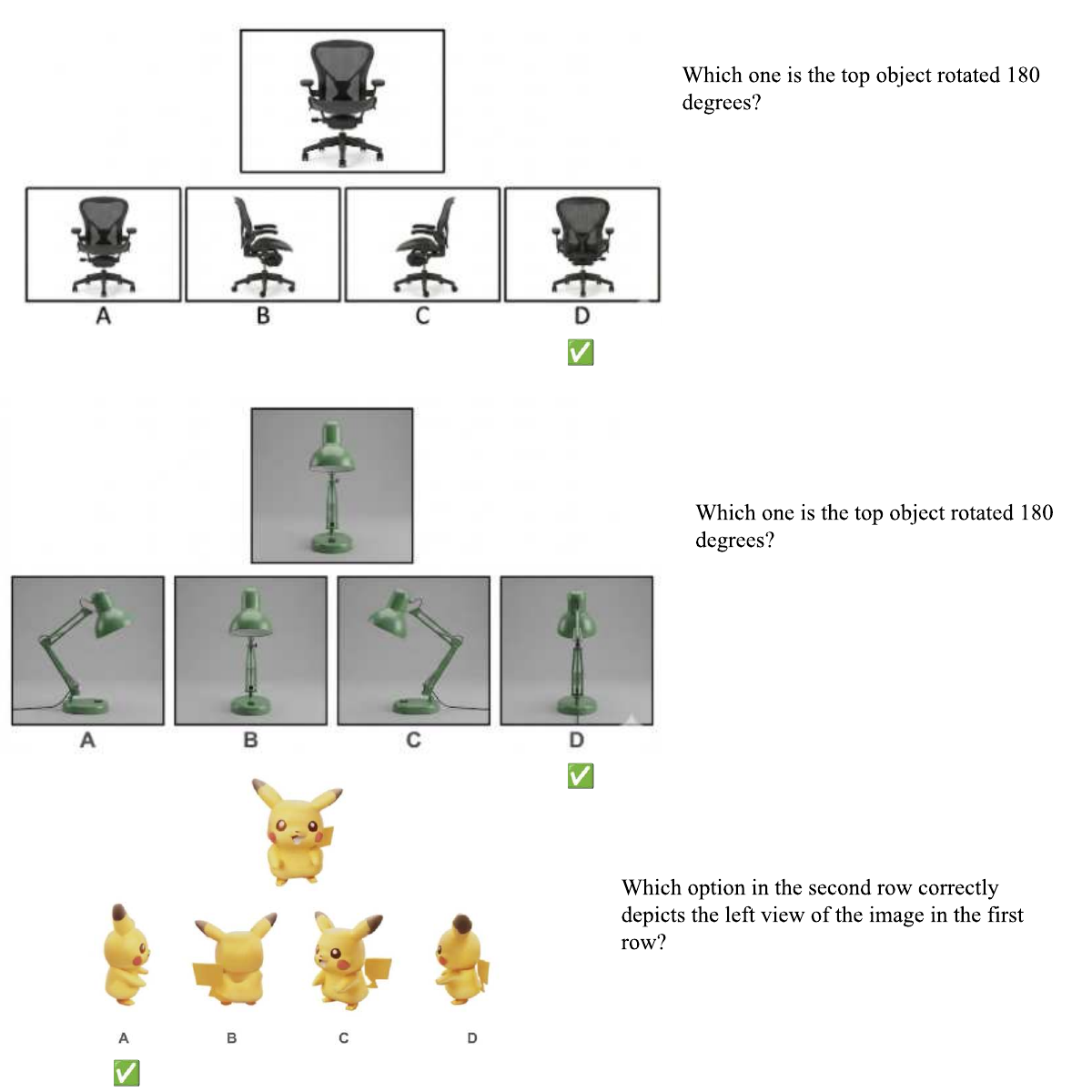}
    \caption{Sample VRIQ natural 3D Visualization items. Each puzzle requires identifying an alternate viewpoint of a real-world object.}
    \label{fig:natural-3d-samples}
\end{figure*}

\begin{figure*}[p]
    \centering
    \includegraphics[width=0.8\linewidth,height=0.85\textheight,keepaspectratio]{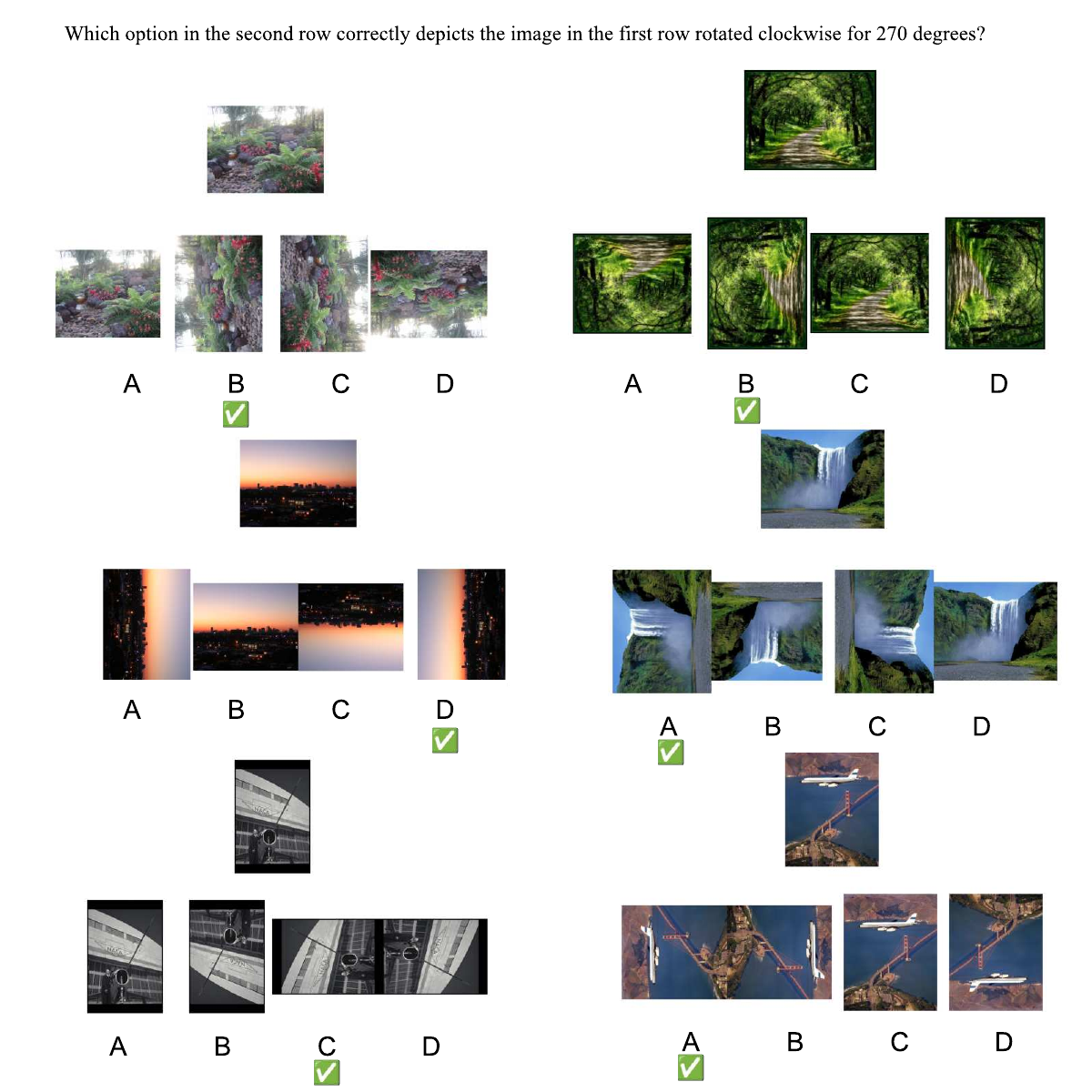}
    \caption{Sample VRIQ natural Figure Rotation items. Each puzzle requires identifying a rotated version of a real-world scene.}
    \label{fig:natural-fr-samples}
\end{figure*}

\begin{figure*}[p]
    \centering
    \includegraphics[width=0.9\linewidth,height=0.85\textheight,keepaspectratio]{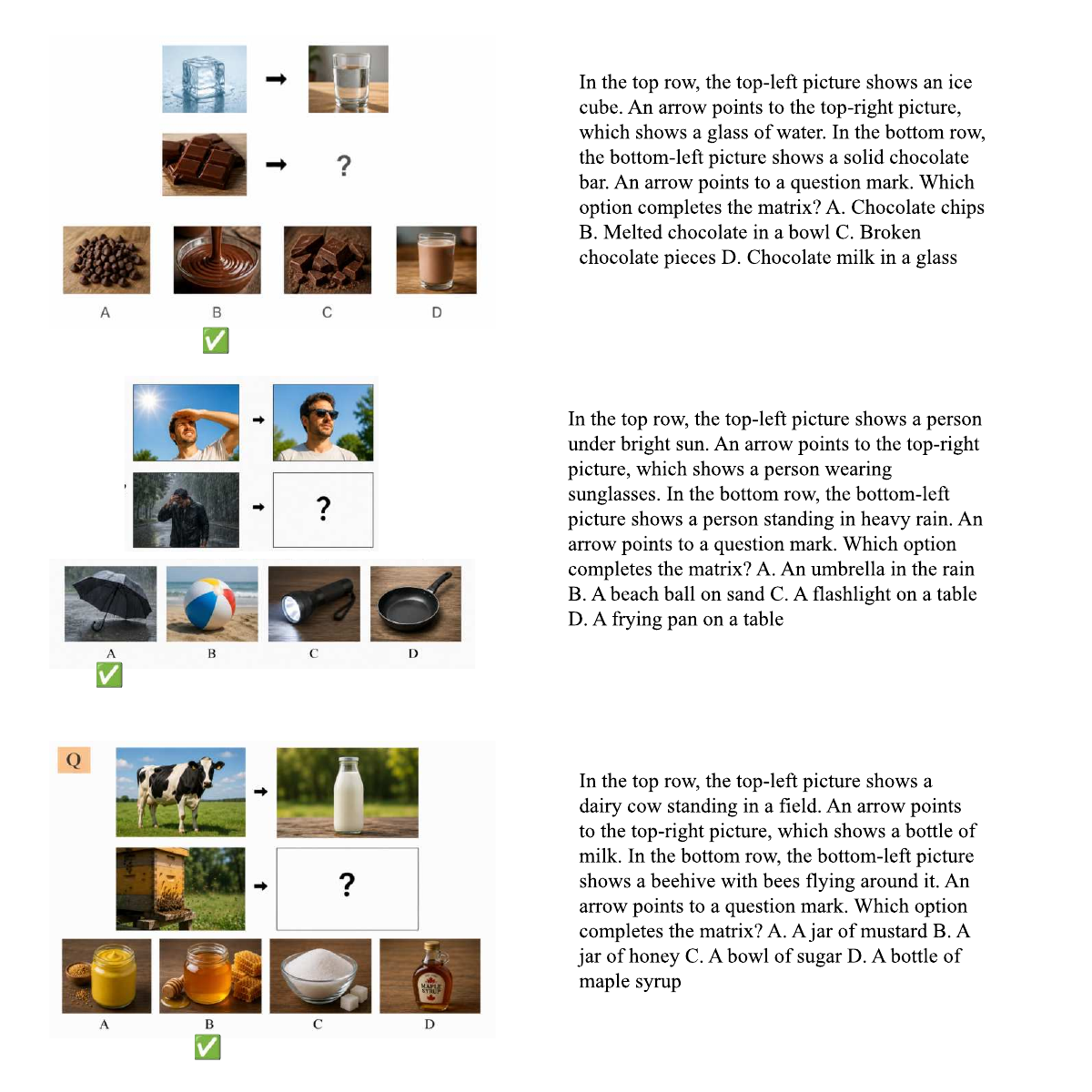}
    \caption{Three representative DiagVRIQ Matrix Prediction items paired with their human-verified ground-truth descriptions.}
    \label{fig:diagvriq-examples}
\end{figure*}

\end{document}